\documentclass[preprint,12pt]{elsarticle}  
\biboptions{numbers,sort&compress}
\usepackage{amsmath,amssymb}
\usepackage{graphicx}
\usepackage{makecell}
\graphicspath{{figures/}} 
\usepackage{booktabs}
\usepackage{multirow}
\usepackage{url}
\usepackage{hyperref} 
\usepackage{subfig}   
\usepackage{caption}  
\usepackage{calc}     
\usepackage{adjustbox} 
\newcommand{\revise}[1]{\textcolor{black}{#1}} 
\newcommand{\newrevise}[1]{\sethlcolor{yellow}\hl{#1}} 
\renewcommand{\revise}[1]{#1}
\renewcommand{\newrevise}[1]{#1}
\newlength{\imgwidth}
\journal{}

\begin{document}

\begin{frontmatter}

\title{GSA-YOLO: A High-Efficiency Framework via Structured Sparsity and Adaptive Knowledge Distillation for Real-Time X-ray Security Inspection}

\author{Jiahao Kong}
\affiliation{organization={SDU-ANU Joint Science College},
            addressline={Shandong University},
            city={Weihai},
            postcode={264200},
            country={China}}

\cortext[mycorrespondingauthor]{Corresponding author}
\ead{zrdi9279@outlook.com}

\begin{abstract}
X-ray security inspection requires accurate real-time detection of prohibited items, but existing models often struggle to balance the challenges of severe occlusion, complex clutter, and strict speed requirements. To overcome these challenges, this paper \revise{proposes} GSA-YOLO, a novel lightweight framework built upon the YOLOv8n architecture, specifically engineered to enhance detection robustness and inference efficiency. GSA-YOLO strategically integrates structured sparsity and adaptive knowledge transfer through three core components: Group Lasso (GL) applied to the network neck for robust feature extraction; Sparse Structure Selection (SSS) applied to the detection head for significant model slimming; and an Adaptive Knowledge Distillation (Ada-KD) mechanism for comprehensive accuracy recovery. This integrated approach synergistically enhances feature representation while pruning redundant channels, maximizing model efficiency without sacrificing performance. Rigorous evaluations on the HiXray and PIDray datasets confirm GSA-YOLO's comprehensive capability, achieving a leading inference speed of 189.62 FPS, accompanied by a reduction in computational cost from 8.7G to 8.0G. Crucially, GSA-YOLO secures mAP50:95 results of 0.531 and 0.679 on HiXray and PIDray, demonstrating 2.4\% and 1.8\% improvements over the baseline, respectively. Compared to other models, GSA-YOLO exhibits enhanced accuracy while maintaining computational efficiency, making it a promising solution for practical X-ray security inspection.
\end{abstract}



\end{frontmatter}

\section{Introduction}
\label{sec:introduction}
Security inspection at public transportation hubs and sensitive areas is an essential component of public safety \cite{intro-first}. X-ray baggage screening systems are widely deployed for this purpose, but conventional approaches rely on manual interpretation, which is susceptible to human error, fatigue, and inconsistency, contributing to missed detections \cite{intro-second}. The advancement of computer vision, particularly deep learning, has opened the possibility of more accurate and consistent automated detection of prohibited items \cite{intro-third}.

However, deploying deep learning models in X-ray security inspection presents challenges that differ from those in natural image recognition. Complex backgrounds and occlusion from the random arrangement and overlap of luggage contents hinder reliable feature extraction \cite{intro-fourth}. Prohibited item detection is also a multi-target problem with large variation in object scale, which increases the likelihood of missed detections for small or partially visible items \cite{intro-fifth}. In addition, high-throughput security checkpoints require real-time inference, imposing a practical constraint on model complexity that must be balanced against detection accuracy \cite{intro-sixth}.

To address these difficulties, methods from three research directions have been explored. On the model compression side, structured sparsity techniques such as Group Lasso \cite{wen_learning_2016,li_group_2020} and Sparse Structure Selection \cite{data-driven_2018} have been applied to reduce model complexity, while knowledge distillation \cite{hinton_distilling_2015,zagoruyko_paying_2017,boutros_adadistill_2025} has been used to recover accuracy lost during compression. On the detection architecture side, numerous methods have been proposed specifically for X-ray imagery, including frequency-aware dual-stream transformers \cite{Zhu2024FDTNet}, split-attention anchor-based detectors \cite{Ding2023Foreign}, collaborative knowledge distillation frameworks \cite{Wei2024Cooperative}, hard negative selection methods \cite{Chang2022Detecting}, incremental learning for occluded instances \cite{Hassan2021Novel}, feature fusion and edge-aware modules \cite{Ma2022EAODNet,Wei2021CFPANet}, and recent attention-enhanced YOLO variants \cite{raj_enhancing_2025,zhou_ei-yolo_2025,wang_improved_2024}. While these methods have advanced specific aspects of X-ray detection, they generally do not simultaneously address detection robustness under complex conditions and the inference efficiency required for real-time deployment.

\revise{A key difficulty lies in the inherent tension among these challenges: designs that improve robustness tend to increase model complexity, while compression techniques that improve inference speed often degrade accuracy, particularly for subtle or occluded targets.} \newrevise{In X-ray security inspection, this tension is particularly evident: prohibited items exhibit overlapping layouts, low contrast, and high material diversity, making fine-grained feature representations important and sensitive to structural compression. Furthermore, while knowledge distillation can partially mitigate accuracy degradation, it introduces additional training stages and hyperparameter sensitivity.} \revise{Most existing methods address only one or two of these aspects without jointly optimizing all three.} To overcome these limitations, we propose GSA-YOLO, a novel lightweight framework based on the YOLOv8n architecture, that strategically integrates structured sparsity and knowledge transfer to simultaneously improve detection robustness and inference efficiency. \revise{Each component is designed to address a specific aspect of this tension: GL applies soft regularization in the neck to preserve feature quality under occlusion; SSS performs structured pruning in the detection head to reduce computational cost; and Ada-KD mitigates the accuracy degradation caused by structural compression.} The main contributions of the proposed GSA-YOLO method are summarized as follows:
\begin{enumerate}
    \item \revise{We propose an integrated sparsity framework that combines Group Lasso (GL) soft regularization in the network neck with Sparse Structure Selection (SSS) hard pruning in the detection head. This dual-stage approach systematically reduces model complexity while maintaining essential feature representations, leading to measurable improvements in FPS and computational efficiency.}
    \item We employ an Adaptive Knowledge Distillation (Ada-KD) mechanism to mitigate the accuracy loss introduced during the sparsity training stages. By adaptively transferring knowledge from a larger pretrained teacher model, Ada-KD facilitates accuracy recovery in the compressed student model.
    \item Experiments on the HiXray and PIDray datasets demonstrate that GSA-YOLO achieves a favorable accuracy-efficiency trade-off, maintaining competitive detection performance while improving inference speed compared to recent YOLO variants.
\end{enumerate}

The remainder of this paper is organized as follows. Section \ref{sec:relatedwork} reviews the relevant research, specifically summarizing the current state of network sparsity, knowledge distillation, and X-ray object detection architectures. Section \ref{sec:preliminaries} introduces the YOLOv8n architecture and the overall GSA-YOLO pipeline as preliminaries. Section \ref{sec:methodology} then presents the technical details of the proposed GSA-YOLO framework. In Section \ref{sec:experiment}, we conduct comprehensive experiments to demonstrate GSA-YOLO's competitive performance on different aspects against mainstream models. Section \ref{sec:conclusion} concludes this paper by summarizing our research, discussing limitations, and suggesting future research directions.

\section{Related Work}
\label{sec:relatedwork}
\subsection{Advances in Network Pruning and Structured Sparsity}
\label{subsec:methodA}
The continuous expansion of deep neural networks necessitates effective model compression techniques for efficient deployment on resource-limited security inspection systems. The foundational concept of network pruning can be traced back to Optimal Brain Damage (OBD) \cite{lecun_optimal_1989}
and later popularized by Deep Compression \cite{han_deep_2016}. However, this unstructured pruning offered limited practical acceleration due to the resulting irregular sparse matrix structure. To realize actual inference speedup, research shifted toward structured sparsity methods, which remove contiguous groups of parameters on entire channels or filters, thereby preserving the dense matrix structure compatible with parallel hardware. Network Slimming \cite{liu_learning_2017} pioneered channel pruning. To achieve more explicit structure removal, the principle of Group Lasso (GL), which enforces group-wise sparsity using a $\ell_2/\ell_1$ mixed-norm penalty \cite{yuan_model_2006}, was adapted for deep neural networks \cite{wen_learning_2016,bui_structured_2021,li_group_2020,shao_mobileprune_2022}. GL and its non-convex variants are generally viewed as soft sparsity methods, excellent for subtle feature refinement, but are often incompatible with parallel hardware for practical inference acceleration, making them less suitable for hard real-time deployment. For high-level structural removal, hard sparsity techniques that directly remove network structures are required. Sparse Structure Selection (SSS) \cite{data-driven_2018} formulates pruning as an explicit optimization problem to select the most critical sparse structures, allowing for direct and significant reduction of GFLOPs and parameters \cite{park_reprune_2024,lin_channel_2020}. However, relying solely on hard sparsity is also proven insufficient: while hard sparsity minimizes model size, it frequently struggles to maintain detection stability and robustness against subtle or complex features, highlighting the necessity for a strategic, multi-stage application that balances compression and fidelity. \newrevise{This trade-off between compression and accuracy motivates the use of knowledge distillation as a recovery mechanism, which we review next.}

\subsection{Knowledge Distillation and Adaptive Knowledge Transfer}
\label{subsec:methodB}
Despite the gains in efficiency from network pruning, deep compression and structural changes inevitably lead to a degradation in model accuracy and robustness. To mitigate this critical performance drop, Knowledge Distillation (KD) \cite{hinton_distilling_2015} has become an indispensable technique. KD transfers generalized knowledge from a high-performing teacher model to a compressed student model \cite{zagoruyko_paying_2017}. This technique is widely employed to recover lost precision for sparsified architectures \cite{zhang_cakdp_2024}. While conventional KD uses a fixed loss structure, recent efforts have focused on adaptive knowledge transfer to account for the capacity gap between Teacher and Student. Adaptive Knowledge Distillation (Ada-KD) represents a notable development, introducing methods that dynamically adjust the distilled knowledge complexity based on the Student's learning progression. Boutros et al. \cite{boutros_adadistill_2025} proposed an AdaDistill approach that dynamically weighs knowledge transfer. \revise{Recent works have explored cross-modal knowledge distillation} \cite{song_cmkd_2025} \revise{and symmetrical learning approaches} \cite{song_symmetrical_2025} \revise{to address data sparsity and asymmetry challenges in knowledge transfer.} \revise{Furthermore, iterative contextual and adaptive strategies have shown promise in image analysis tasks} \cite{umirzakova_iterative_2025}. \revise{Additionally, advanced attention mechanisms including frequency domain learning, dual meta attention, and meta PID networks have demonstrated effectiveness in handling complex image denoising and enhancement tasks} \cite{ma_learning_2024,ma_flexible_2023,ma_meta_2022}. This adaptive paradigm is well-suited for highly sparse models, as it offers a more flexible mechanism for feature recovery compared to fixed-loss KD. However, the effectiveness of integrating such advanced Ada-KD techniques with multi-stage sparse architectures remains an underexplored challenge in object detection literature. \newrevise{Moreover, existing KD methods are largely validated on general detection benchmarks and do not account for the specific challenges of X-ray security inspection, where occlusion and low contrast impose additional demands on feature preservation, as reviewed in the following subsection.}

\subsection{Object Detection Architectures in X-ray Security Imaging}
\label{subsec:methodC}
X-ray security inspection is a safety-critical application facing distinct and severe challenges, primarily severe object occlusion and high background clutter. The development of specialized datasets, such as SIXray \cite{miao_sixray_2019}, HiXray \cite{tao_towards_2021}, and the challenging PIDray dataset \cite{wang_towards_2021} with its explicitly hidden test sets, has highlighted the need for robust detection under extreme interference. The architectural evolution in this field has been driven by the push for real-time performance. This began with slower two-stage detectors like Faster R-CNN \cite{ren_faster_2015}, but the push for speed quickly favored highly efficient one-stage detectors. You Only Look Once (YOLO) \cite{redmon_you_2016} revolutionized this paradigm, and its variants, including YOLOv3 \cite{redmon_yolov3_2018}, continuously refined efficiency. A major breakthrough in one-stage detection came with RetinaNet \cite{lin_focal_2017}, which introduced Focal Loss to successfully address the extreme foreground-background class imbalance inherent in dense prediction, boosting one-stage detectors' accuracy to parity with two-stage methods. Subsequent one-stage models like YOLOv5 \cite{jocher_ultralyticsyolov5_2020} and the state-of-the-art YOLOv8 \cite{ultralytics_yolov8} have further refined efficiency through innovations like multi-scale feature pyramids and anchor-free, decoupled-head designs. \revise{Dual attention mechanisms have also been explored for context-aware feature learning in agricultural detection applications} \cite{zhou_dual_2026}. \revise{Recent advances have explored alternative architectures for achieving efficiency-accuracy trade-offs, such as Mamba-based frameworks that achieve real-time performance without requiring post-hoc compression} \cite{luo_epdd_2025}. Existing X-ray detection methods typically build on these architectures, prioritizing accuracy enhancements through architectural improvements like attention mechanisms \cite{raj_enhancing_2025,zhou_ei-yolo_2025} or refined feature fusion networks \cite{noauthor_scanguard-yolo_nodate,wang_pad-f_2024,han_sc-yolov8_2023,cani_illicit_2025}, often sacrificing the strict real-time efficiency required for mass deployment. We select the highly efficient YOLOv8 architecture as our core baseline.

\begin{table}[htbp]
\centering
\caption{\revise{Summary of representative related works across three research directions.}}\label{tab:related_summary}
\resizebox{\textwidth}{!}{
\begin{tabular}{l|l|l|l}
\hline
\revise{Direction} & \revise{Method} & \revise{Core Contribution} & \revise{Limitation} \\
\hline
\revise{Structured Sparsity} & \revise{Network Slimming} & \revise{Channel-level pruning via BN scaling factors} & \revise{Uniform application without position awareness} \\
\revise{Structured Sparsity} & \revise{Group Lasso} & \revise{Soft group-wise sparsity regularization} & \revise{Incompatible with hard real-time deployment alone} \\
\revise{Structured Sparsity} & \revise{SSS} & \revise{Explicit structure selection via learnable scaling} & \revise{Accuracy drop requires post-pruning recovery} \\
\revise{Knowledge Distillation} & \revise{Hinton KD} & \revise{Soft-label knowledge transfer via temperature scaling} & \revise{Fixed loss weight, no adaptation to student progress} \\
\revise{Knowledge Distillation} & \revise{AdaDistill} & \revise{Dynamic adaptive weighting of distillation} & \revise{Not validated on structurally compressed models} \\
\revise{Knowledge Distillation} & \revise{Cross-modal KD} & \revise{Multi-modal knowledge transfer for data sparsity} & \revise{Domain-specific, not generalized to pruned architectures} \\
\revise{X-ray Detection} & \revise{Faster R-CNN} & \revise{Two-stage accurate detection} & \revise{Insufficient inference speed for real-time deployment} \\
\revise{X-ray Detection} & \revise{YOLOv8} & \revise{Anchor-free decoupled head, high efficiency} & \revise{Not optimized for X-ray occlusion and clutter} \\
\revise{X-ray Detection} & \revise{ESI-YOLO / GEMA-YOLO} & \revise{X-ray specific architectural enhancements} & \revise{No structural compression for deployment efficiency} \\
\hline
\end{tabular}
}
\end{table}

\revise{The three research directions reviewed above share a common underlying tension that has not been jointly addressed. Structured sparsity methods (GL, SSS) can reduce model size and FLOPs, but hard pruning alone causes accuracy degradation that is particularly severe for subtle prohibited items in X-ray imagery. Knowledge distillation can recover this accuracy loss, yet existing adaptive KD strategies have not been systematically validated in combination with multi-stage structured sparsity pipelines. Meanwhile, X-ray detection architectures prioritize robustness against occlusion and clutter, but none of the existing lightweight detectors simultaneously address compression, accuracy recovery, and domain-specific robustness in a unified framework. The intersection of these three gaps defines the core scientific problem: how to achieve aggressive structural compression while maintaining detection robustness in X-ray security inspection scenarios.}

The preceding review highlights a fundamental gap in achieving deployable, high-speed X-ray detection models. Current pruning methods fail to strike the necessary balance between extreme compression and maintaining feature robustness, leading to unacceptable accuracy loss on subtle prohibited items. Furthermore, while advanced KD techniques exist, their systematic integration and validation for recovering performance in models subjected to aggressive, structured sparsity remains unexplored. To address this efficiency-robustness gap, we propose a novel Sparse-Constraint YOLO Detection Framework, GSA-YOLO. This framework integrates a hybrid structured sparsity approach with an Ada-KD mechanism, targeting real-time, accurate X-ray security detection.

\begin{figure}[htbp]
    \centering
    \includegraphics[width=1\textwidth]{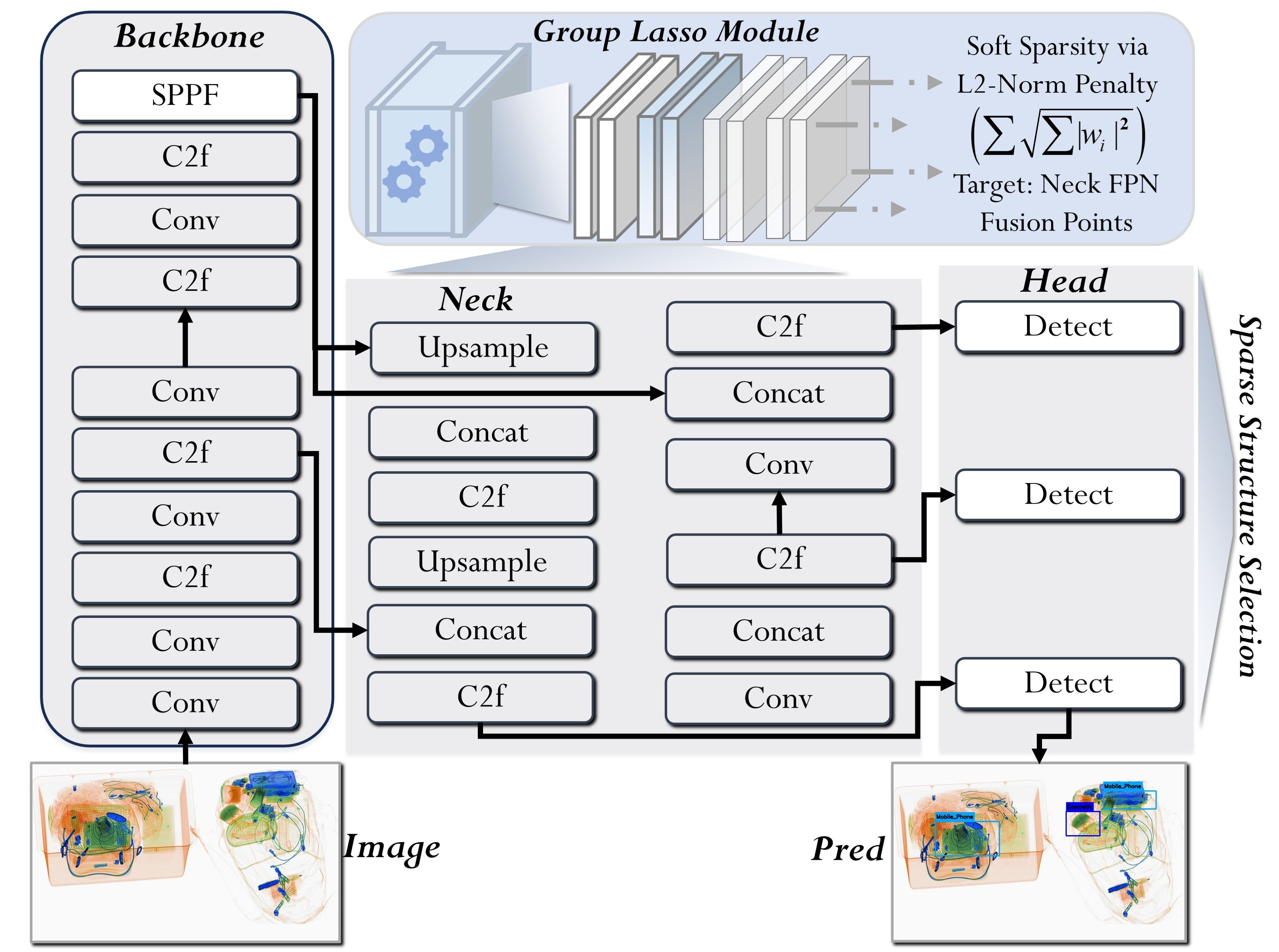} 
    \caption{\revise{GSA-YOLO framework overview based on YOLOv8n for X-ray security inspection. The architecture consists of three main components: Backbone (Conv, C2f, SPPF modules for basic visual feature extraction), Neck (Upsample, Concat, C2f, Conv modules implementing multi-scale feature fusion with FPN structure, targeted by Group Lasso), and Head (three Detect branches for multi-scale object detection, targeted by Sparse Structure Selection).}}\label{fig:framework} 
\end{figure}

\section{Preliminaries}
\label{sec:preliminaries}
\newrevise{GSA-YOLO is built upon the YOLOv8n architecture}~\cite{ultralytics_yolov8}\newrevise{, a compact anchor-free detector comprising three components: a Backbone (Conv, C2f, SPPF) for feature extraction, a Neck (FPN-based multi-scale fusion) for feature aggregation, and a decoupled detection Head with three scale-specific prediction branches (P3, P4, P5). YOLOv8n achieves a strong speed-accuracy baseline at 8.7G GFLOPs and 6.2MB model size, making it a suitable foundation for structured compression. The three-stage GSA-YOLO pipeline applies Group Lasso (GL) regularization to the Neck, Sparse Structure Selection (SSS) pruning to the Head, and Adaptive Knowledge Distillation (Ada-KD) fine-tuning to recover accuracy, as illustrated in Figure}~\ref{fig:framework}\newrevise{.}

\section{Proposed Method}
\label{sec:methodology}
This section details the three proposed components of GSA-YOLO: Group Lasso (GL), Sparse Structure Selection (SSS), and Adaptive Knowledge Distillation (Ada-KD). The framework employs a multi-stage optimization process divided into two cascaded layers: the Sparsity-Compression Layer, where GL first establishes a differentiable soft-sparsity foundation and SSS subsequently executes the irreversible structural compression; and the Accuracy-Recovery Layer, where Ada-KD fine-tunes the compressed model to restore performance. This sequential approach is necessary because SSS is a non-differentiable operation, preventing simultaneous optimization of all components. \revise{The strategic application of different sparsity techniques to distinct architectural components is theoretically motivated by their functional roles, as shown in} Table~\ref{tab:layer_strategy}\revise{, where GL targets the neck's feature fusion layers where multi-scale information integration creates channel redundancy, while SSS targets the detection head where hard pruning can eliminate redundant channels without compromising detection performance.} \revise{Throughout this paper, ``detection head'' refers to the overall detection module (model.22) as a single architectural unit, while ``detection heads'' refers to the three scale-specific prediction branches (P3, P4, P5) contained within it. The complete mapping between YOLOv8n model layer indices and their architectural roles is provided in Table}~\ref{tab:layer_strategy}\revise{, where layers model.12--21 correspond to the Neck and model.22 corresponds to the detection head.}

\begin{table}[htbp]
\centering
\caption{\revise{Strategic application of sparsity techniques across YOLOv8n architecture}}\label{tab:layer_strategy}
\resizebox{\textwidth}{!}{
\begin{tabular}{c|l|l|l}
\hline
\revise{Layer} & \revise{Module} & \revise{Function} & \revise{Sparsity Strategy} \\
\hline
\revise{0-9} & \revise{Backbone} & \revise{Feature extraction} & \revise{No modification} \\
\revise{10-11} & \revise{Neck (Upsample + Concat)} & \revise{Multi-scale fusion} & \revise{-} \\
\revise{12} & \revise{Neck (C2f)} & \revise{Feature refinement} & \revise{\textbf{GL applied}} \\
\revise{13-14} & \revise{Neck (Upsample + Concat)} & \revise{Multi-scale fusion} & \revise{-} \\
\revise{15} & \revise{Neck (C2f)} & \revise{Feature refinement} & \revise{\textbf{GL applied}} \\
\revise{16} & \revise{Neck (Conv)} & \revise{Downsampling} & \revise{\textbf{GL applied}} \\
\revise{17} & \revise{Neck (Concat)} & \revise{Feature fusion} & \revise{-} \\
\revise{18} & \revise{Neck (C2f)} & \revise{Feature refinement} & \revise{\textbf{GL applied}} \\
\revise{19} & \revise{Neck (Conv)} & \revise{Downsampling} & \revise{\textbf{GL applied}} \\
\revise{20} & \revise{Neck (Concat)} & \revise{Feature fusion} & \revise{-} \\
\revise{21} & \revise{Neck (C2f)} & \revise{Feature refinement} & \revise{\textbf{GL applied}} \\
\revise{22} & \revise{Head (Detect)} & \revise{Multi-scale detection} & \revise{\textbf{SSS applied}} \\
\hline
\end{tabular}
}
\end{table}

\revise{To further quantify the rationale behind this allocation, we analyze the channel redundancy of each layer group. As shown in Table}~\ref{tab:redundancy_analysis}\revise{, Head layers consistently exhibit higher redundancy scores than Neck layers, empirically supporting the application of hard SSS pruning to the Head and soft GL regularization to the Neck.}

\begin{table}[htbp]
\centering
\caption{\revise{Channel redundancy analysis of Neck and Head layers in YOLOv8n, supporting the design rationale for GL-Neck and SSS-Head allocation.}}\label{tab:redundancy_analysis}
\resizebox{\textwidth}{!}{
\begin{tabular}{c|l|c|c|c}
\hline
\revise{Group} & \revise{Layer} & \revise{Channels} & \revise{High Corr Ratio} & \revise{Redundancy Score} \\
\hline
\revise{Head} & \revise{head.box.cv2.0} & \revise{64} & \revise{0.36} & \revise{0.42} \\
\revise{Head} & \revise{head.box.cv2.1} & \revise{64} & \revise{0.34} & \revise{0.42} \\
\revise{Head} & \revise{head.box.cv2.2} & \revise{64} & \revise{0.31} & \revise{0.40} \\
\revise{Head} & \revise{head.cls.cv3.0} & \revise{8}  & \revise{0.33} & \revise{0.41} \\
\revise{Head} & \revise{head.cls.cv3.1} & \revise{8}  & \revise{0.32} & \revise{0.41} \\
\revise{Head} & \revise{head.cls.cv3.2} & \revise{8}  & \revise{0.32} & \revise{0.41} \\
\revise{Neck} & \revise{neck.model.16}  & \revise{64} & \revise{0.18} & \revise{0.24} \\
\revise{Neck} & \revise{neck.model.17}  & \revise{192}& \revise{0.18} & \revise{0.23} \\
\revise{Neck} & \revise{neck.model.18}  & \revise{128}& \revise{0.17} & \revise{0.23} \\
\revise{Neck} & \revise{neck.model.19}  & \revise{128}& \revise{0.18} & \revise{0.24} \\
\revise{Neck} & \revise{neck.model.20}  & \revise{384}& \revise{0.18} & \revise{0.24} \\
\revise{Neck} & \revise{neck.model.21}  & \revise{256}& \revise{0.17} & \revise{0.23} \\
\hline
\end{tabular}
}
\end{table}

\subsection{Group Lasso}
\label{subsec:GL}
GL is strategically applied to the Neck part of the YOLOv8 model, \revise{specifically targeting the key convolutional layers model.12, model.15, model.16, model.18, model.19, and model.21, which correspond to the main feature fusion points in the Feature Pyramid Network}. \revise{As a training-time feature refinement mechanism, GL does not directly remove channels but instead identifies and suppresses less important feature channels through soft regularization, preparing the network for subsequent structured pruning.} GL is a regularization technique that enforces group-wise sparsity, which is crucial for achieving channel-level sparsity in neural networks. The objective is to shrink the weights of entire groups, effectively identifying and eliminating redundancy in the feature extraction process. Unlike unstructured pruning, which operates on individual weights, GL applies the penalty to a group of weights simultaneously. In our implementation, we treat the entire weight vector of each output channel as a single group. The loss function for a single convolutional layer is formulated as:

\begin{equation}
L_{GL}^{\left( l \right)}=L_{det}+\beta \sum_{g=1}^{G_l}{||\mathbf{w}_{g}^{\left( l \right)}||_{2}} 
\end{equation}

The total loss for the entire model is the global detection loss plus the sum of GL regularization terms across all neck layers:

\begin{equation}
\label{eq:total_GL_loss}
L_{\text{total\_GL}}=L_{det}+\beta \sum_{l=1}^L \sum_{g=1}^{G_l}{||\mathbf{w}_{g}^{\left( l \right)}||_{2}}
\end{equation}
where $L_{det}$ represents the global base detection loss computed once; $\mathbf{w}_{g}^{\left( l \right)}$ is the entire weight vector of the $g$-th output channel of the $l$-th convolutional layer, which is treated as a group; \revise{$G_l$ is the number of output channels in the $l$-th layer}; and $\beta$ is the regularization hyperparameter that controls the strength of the penalty for Group Lasso. The $ \ell _2$-norm $ ||\mathbf{w}_{g}^{\left( l \right)}||_{2} $ ensures that if a channel's weights are shrunk, they are shrunk together, thereby encouraging channel-wise sparsity. $L$ represents the total number of convolutional layers in the neck part of the YOLOv8n model.

The gradient component contributed by the Group Lasso penalty term with respect to $\mathbf{w}_{g}^{\left( l \right)} $ is calculated as shown in Equation (\ref{eq:wg_gradient}). This penalty gradient is added to the gradient of the base detection loss $L_{det}^{\left( l \right)}$ during optimization, pushing the weights of less important channels toward zero, thereby enabling channel-wise sparsity. The selection of the regularization hyperparameter $\beta$ is critical for balancing model performance and the degree of induced sparsity; we set the default value of $\beta$ to 1e-4.
\begin{equation}
\label{eq:wg_gradient}
\nabla _{\mathbf{w}_{g}^{\left( l \right)}}\left( \beta ||\mathbf{w}_{g}^{\left( l \right)}||_{2} \right) =\beta \frac{\mathbf{w}_{g}^{\left( l \right)}}{\left\| \mathbf{w}_{g}^{\left( l \right)} \right\| _2 + \varepsilon}
\end{equation}
where \revise{$\varepsilon$ is a small positive constant (typically $10^{-8}$) added to prevent division by zero when $\mathbf{w}_{g}^{\left( l \right)}$ is a zero vector.}

\subsection{Sparse Structure Selection}
\label{subsec:SSS}
SSS is the hard pruning module applied after GL, \revise{specifically targeting the detection head (model.22) which contains the convolutional layers for the three different scale detection outputs in YOLOv8n's multi-scale detection architecture}. These layers in the detection head are checked to exhibit greater channel redundancy, presenting a prime opportunity for aggressive hard pruning. The primary goal of SSS is to achieve model slimming and computational efficiency by eliminating less important channels, based on the learned channel scaling factors $\lambda$ during the pretraining process, while preserving essential feature representations.

SSS employs $\ell_1$-norm regularization on the channel scaling factors $\lambda$, which are designed to scale the outputs of specific structures, which is the channels in our module. The loss function for a single convolutional layer $l$ is formulated as:
\begin{equation}
L_{SSS}^{\left( l \right)}=L_{det}+\gamma \sum_{c=1}^{C_l}{||\lambda _{l,c}||_1} 
\end{equation}

The total loss for the entire model across all layers is:
\begin{equation}
L_{\text{total\_SSS}}=L_{det}+\gamma \sum_{l=1}^L \sum_{c=1}^{C_l}{||\lambda _{l,c}||_1}
\end{equation}
where \revise{$L_{det}$ represents the global base detection loss computed once};   $\lambda _{l,c}$ is the dedicated scaling factor for the $c$-th output channel in the $l$-th layer; $C_l$ is the total number of output channels in the $l$-th layer; $||\lambda _{l,c}||_1$ is the $\ell_1$-norm of the scaling factor, which promotes channel sparsity by forcing less important factors toward zero; $\gamma$ is the regularization hyperparameter that controls the strength of the sparsity penalty for SSS; and $L$ represents the total number of convolutional layers targeted for pruning in the head part of the YOLOv8 model, which is exactly 6.

After pretraining, we determine a pruning threshold $\tau$ based on the distribution of the scaling factors $ \lambda _{l,c} $. \revise{Specifically, given a target pruning ratio, we collect all scaling factors from the detection head branches, sort them by magnitude, and set $\tau$ to the corresponding percentile value, ensuring that channels with $|\lambda_{l,c}| < \tau$ are pruned while preserving a minimum number of channels for network functionality. The default value is set to $\tau = 0.001$, which consistently achieves the best detection performance on both datasets as validated by sensitivity analysis in Section}~\ref{subsec:hyperparameter sensitivity analysis}. Channels whose scaling factors $ \lambda _{l,c}$ are below the threshold $\tau$ are subsequently pruned from the network as Equation (\ref{eq:lambda_threshold}) illustrates. After pruning, the model undergoes fine-tuning to recover performance loss caused by the pruning process. The gradient of the loss with respect to $ \lambda _{l,c} $ is shown in Equation (\ref{eq:gradient_lambda}).
\begin{equation}
\label{eq:lambda_threshold}
\lambda _{l,c}=\begin{cases} 	0&		\text{if} ||\lambda _{l,c}||_1<\tau\\ 	\lambda _{l,c}&		\text{otherwise}\\ \end{cases} 
\end{equation}
\begin{equation}
\label{eq:gradient_lambda}
\partial_{\lambda_{l,c}}\left(\gamma|\lambda_{l,c}|\right) = \begin{cases} \gamma\operatorname{sign}(\lambda_{l,c}) & \text{if } \lambda_{l,c}\neq 0 \\ 0 & \text{if } \lambda_{l,c}=0 \end{cases}
\end{equation}
where \revise{$0$ is chosen as the subgradient at $\lambda_{l,c}=0$, which is a valid selection within the subgradient range $[-\gamma,\, \gamma]$ and is standard in proximal gradient optimization.}

\begin{figure}[htbp]
\centering
\includegraphics[width=1\textwidth]{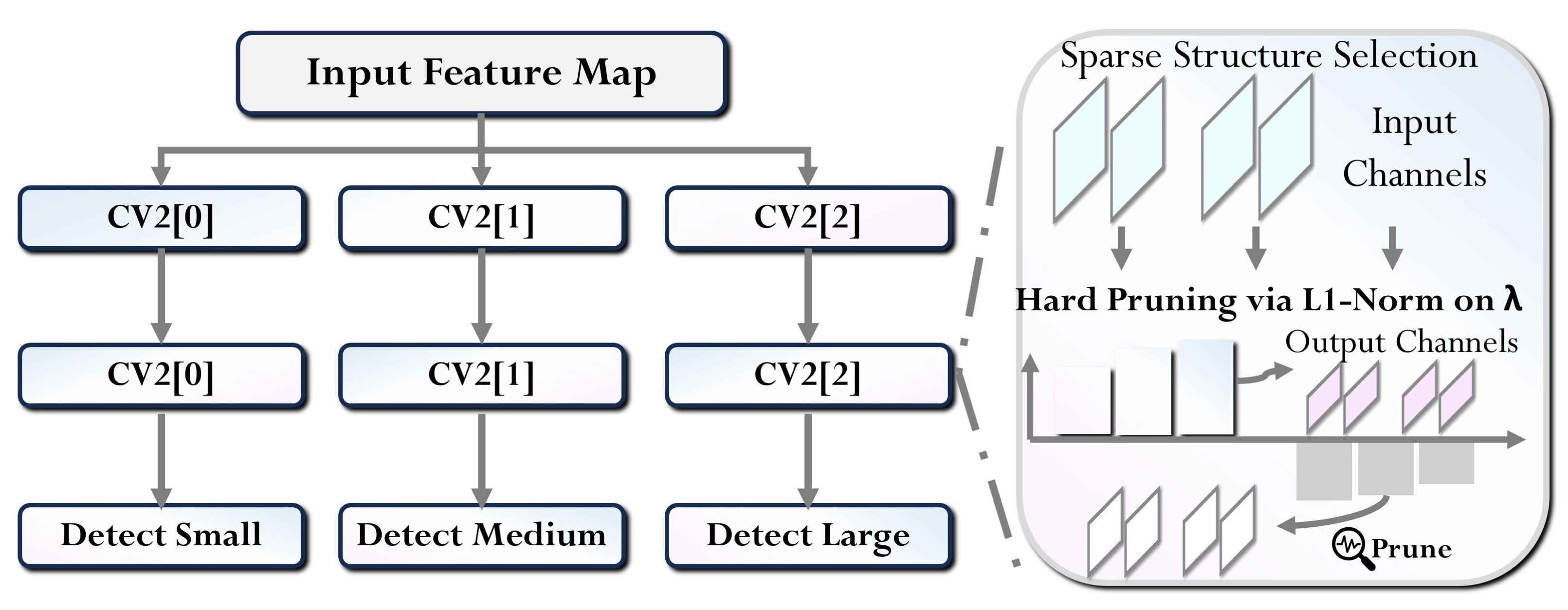}
\caption{The detailed illustration of SSS module. SSS performs channel pruning based on the $\ell_1$ regularized channel scaling factor $\lambda$. Gray channels whose scaling factors $\lambda < \tau$ are inactive and can be pruned.}\label{fig:sss_details}
\end{figure}

The local details of the SSS pruning process, applied to the convolutional layers in \text{model.22} Detection Head, are illustrated in Figure \ref{fig:sss_details}. The choice of $\gamma$ is important for balancing the trade-off between model accuracy and speed, whose default value is set to 1e-3. \revise{Rather than applying $\gamma$ at full strength throughout training, we adopt a warmup-and-ramp schedule: sparsity regularization is introduced after an initial warmup of 10 epochs and then linearly increased to the target value $\gamma = 1\times10^{-4}$ over the subsequent 40 epochs, which stabilizes early training and avoids premature channel suppression.}

\subsection{Adaptive Knowledge Distillation}
\label{subsec:ada-kd}
\begin{figure}[htbp]
\centering
\includegraphics[width=1\textwidth]{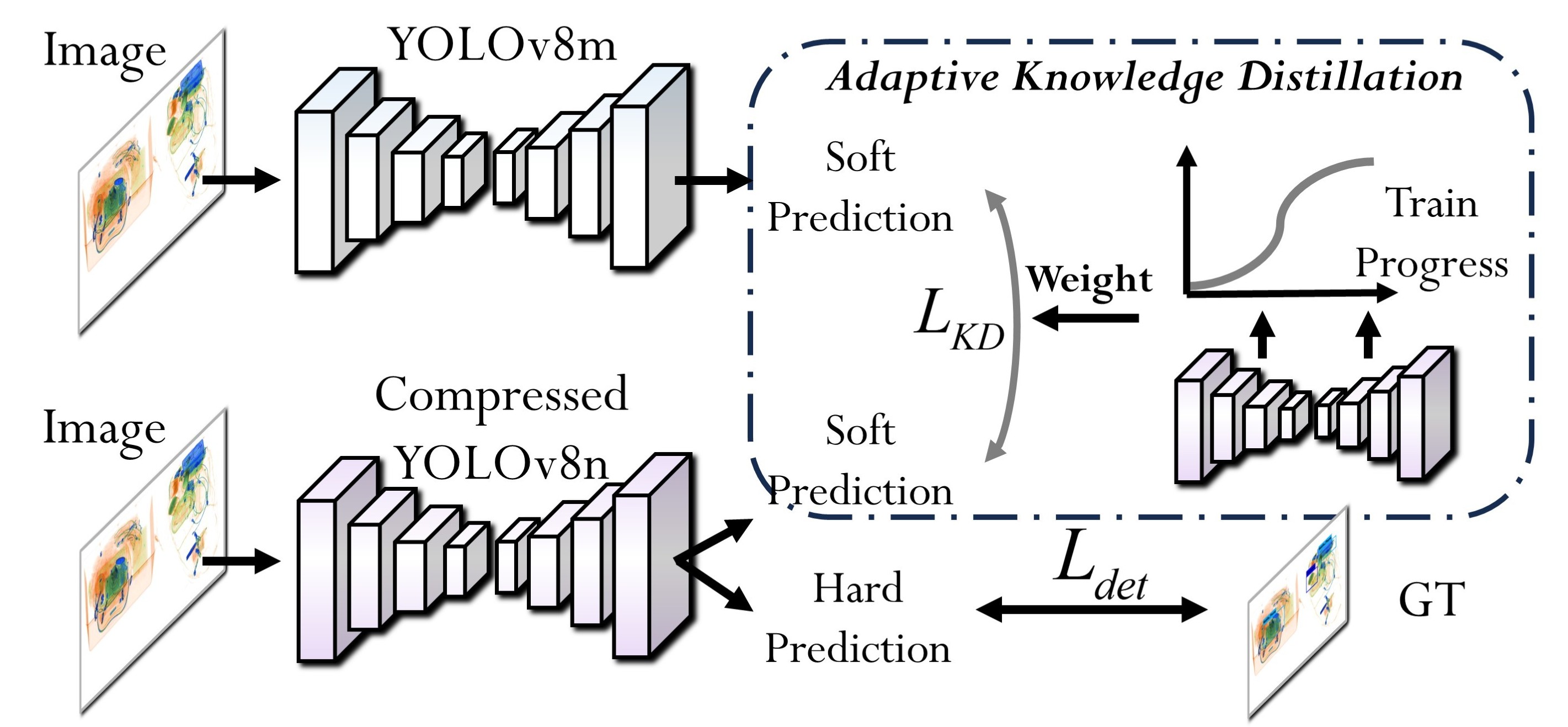}
\caption{The detailed illustration of Ada-KD mechanism.}\label{fig:Ada-KD_details}
\end{figure}

Ada-KD is employed as the final stage to effectively recover the performance of the compressed YOLOv8n model after the application of GL and SSS. The core principle of Ada-KD is to transfer valuable knowledge from a pretrained and high-performing teacher model, the pretrained YOLOv8m model after the comparison in Section \ref{subsec:comparison experiment}, to the sparse student model via a progress-based adaptive weighting mechanism. \revise{The teacher model (YOLOv8m) remains unmodified and retains its full channel structure without any GL or SSS modifications. Since our distillation operates on the output-level classification probability distributions rather than intermediate feature maps, both teacher and student produce probability vectors of identical dimensionality $C$ (the number of classes) after softmax, regardless of the internal channel differences caused by pruning in the student's detection head.} This mechanism dynamically adjusts the influence of the distillation loss based on the student's learning progress, ensuring a balanced optimization between learning from the teacher's soft labels and the ground-truth hard labels. The detailed illustration of Ada-KD mechanism is shown in Figure \ref{fig:Ada-KD_details}.

The total loss function used during the Ada-KD phase is defined as:
\begin{equation}
\label{eq:total_ada}
L_{\text{total\_Ada}}=L_{det}+\lambda _0 \cdot \alpha \left( t \right) L_{KD}
\end{equation}
where $L_{\text{total\_Ada}}$ represents the overall loss during the distillation phase. $L_{det}$ is the base detection loss calculated using the ground-truth hard labels; $L_{KD}$ is the knowledge distillation loss, which is computed using the soft labels provided by the teacher model; $\lambda _0$ is a hyperparameter defining the maximum influence of the distillation loss; and $\alpha \left( t \right) $ is the adaptive weight function that dynamically controls the decay rate of the distillation loss, and it is time-dependent.

The adaptive weight function $ \alpha \left( t \right) $ is calculated using the Sigmoid function in Equation (\ref{eq:alpha_t}), which is designed to adaptively decrease the KD influence as the training progresses:
\begin{equation}
\label{eq:alpha_t}
\alpha \left( t \right) =\frac{1}{1+\exp \left( \theta \left( t-T_{\text{mid}} \right) \right)} 
\end{equation}
where $t$ is the normalized training progress, defined as the current epoch divided by the total number of distillation epochs, thus ranging from 0 to 1; $T_{\text{mid}}$ is a constant, set to 0.5, representing the normalized progress point where the decay factor $\alpha(t)$ drops to 0.5; and $\theta$ is the scaling factor that controls the steepness and transition rate of the Sigmoid curve.

\revise{The distillation loss $L_{KD}$ is computed using Kullback-Leibler (KL) divergence} \cite{hinton_distilling_2015}, \revise{which measures the difference between the teacher's and student's predicted probability distributions for both classification and bounding box regression logits. A temperature coefficient $T=2.0$ is applied to soften the probability distributions before computing the KL divergence, controlling the smoothness of the soft labels during knowledge transfer.} The formulation is as follows:
\begin{equation}
L_{KD}=T^2 \cdot D_{KL}\left( Q_{teacher}^{(T)}\,\|\,Q_{student}^{(T)} \right) =T^2\sum_{i=1}^C Q_{teacher}^{(T)}\left( i \right) \log \frac{Q_{teacher}^{(T)}\left( i \right)}{Q_{student}^{(T)}\left( i \right)}
\end{equation}
where $C$ is the number of classes; $T=2.0$ is the temperature coefficient; and $Q_{teacher}^{(T)}(i) = \frac{\exp(z_{teacher,i}/T)}{\sum_j \exp(z_{teacher,j}/T)}$, $Q_{student}^{(T)}(i) = \frac{\exp(z_{student,i}/T)}{\sum_j \exp(z_{student,j}/T)}$ are the temperature-scaled softmax probabilities for class $i$ from the teacher and student model, respectively. The $T^2$ factor compensates for the gradient magnitude reduction caused by temperature scaling. 

By minimizing $L_{KD}$, the student model is encouraged to mimic the fine-grained decision boundaries encoded in the teacher's soft labels, which is crucial for recovering performance and enhancing robustness in complex X-ray imaging scenarios. The scaling factors $\lambda_0$ and $\theta$ primarily control the overall contribution and balance between hard label learning and knowledge transfer throughout the training phase; we set their default values to 4 and 15, respectively. \revise{Rather than applying differential processing across the three detection scales, we adopt a unified distillation strategy that applies the same KL divergence-based loss across all detection heads, prioritizing implementation simplicity and training stability while the adaptive weighting mechanism provides sufficient dynamic control over the distillation process.}

\section{Experiment}
\label{sec:experiment}
\revise{Building upon the GSA-YOLO framework proposed in the previous section, which integrates GL for neck-level feature regularization, SSS for detection head pruning, and Ada-KD for accuracy recovery, this section presents experiments to empirically evaluate each design decision and the overall framework performance.}

\subsection{Experimental Setup}
\label{subsec:experimental setup}
\subsubsection{Datasets}
\label{subsubsec:datasets}
To evaluate our model across challenging real-world X-ray security scenarios, we evaluate our model on two widely used X-ray datasets: HiXray \cite{tao_towards_2021} and PIDray \cite{wang_towards_2021}. The HiXray contains 45,364 X-ray images, with 102,928 labeled instances collected from real-world airport scans. It features 8 categories of prohibited items, providing a rich diversity of complex and occluded scenarios. The PIDray contains 47,677 images across 12 categories, specifically emphasizing the detection of hidden and occluded objects, allowing us to rigorously test the robustness of the model. Both HiXray and PIDray datasets provide independent and dedicated test sets for final assessment, which have 9096 and 18220 images, respectively. Specially, PIDray offers three distinct testing partitions, easy, hard, and hidden at a ratio about 5:2:3, to separately evaluate model's performance under various occlusion and clutter scenarios. We divided the original training partitions of both datasets into new training and validation sets at an 8:2 ratio. The detailed dataset partition and class distribution are presented in Table \ref{tab:class_distribution_combined}.

\begin{table}[ht]
\caption{The detailed dataset partition and class distribution of HiXray and PIDray}\label{tab:class_distribution_combined}
\centering
\resizebox{\textwidth}{!}{
\begin{tabular}{l l l l}
\toprule
Dataset (Total/Train/Val/Test) & Class & Count \\
\midrule
\multirow{8}{*}{\makecell[c]{HiXray (45364/29014/7254/9096)}} 
& Portable\_Charger\_1 & 12421 \\
& Portable\_Charger\_2 & 7788 \\
& Water & 3092 \\
& Laptop & 10042 \\
& Mobile\_Phone & 5383 \\
& Tablet & 4918 \\
& Cosmetic & 9949 \\
& Nonmetallic\_Lighter & 883 \\
\midrule
\multirow{12}{*}{\makecell[c]{PIDray (47677/23566/5891/18220)}} 
& Baton & 2399 \\
& Pliers & 6814 \\
& Hammer & 6299 \\
& Powerbank & 8116 \\
& Scissors & 7060 \\
& Wrench & 6437 \\
& Gun & 3757 \\
& Bullet & 2957 \\
& Sprayer & 4227 \\
& Handcuffs & 3388 \\
& Knife & 5549 \\
& Lighter & 6157 \\
\bottomrule
\end{tabular}
}
\end{table}

\subsubsection{Evaluation Metrics}
Following standard object detection practices, we evaluate the proposed model using the following metrics to assess detection accuracy: Precision (P), Recall (R), mAP50, mAP50:95, and model efficiency: Frames Per Second (FPS) and GFLOPs. The definition of these metrics relies on True Positives (TP), False Positives (FP), and False Negatives (FN). Furthermore, the Intersection over Union (IoU), representing the ratio of intersection and concatenation between the predicted bounding box and the ground truth box, is used to determine if a detection counts as a TP.

Precision (P) assesses the model's accuracy, reflecting the ratio of correctly predicted positive samples to all detected samples, and is calculated as shown in Equation (\ref{eq:precision}):
\begin{equation}
\label{eq:precision}
\text{Precision} = \frac{TP}{TP + FP}
\end{equation}

Recall (R) evaluates the model's completeness, measuring the ratio of correctly predicted positive samples to the total number of actual positive samples, and is calculated as shown in Equation (\ref{eq:recall}):
\begin{equation}
\label{eq:recall}
\text{Recall} = \frac{TP}{TP + FN}
\end{equation}

The Average Precision (AP) for a single class is defined as the area under the Precision-Recall curve, quantifying the detection performance across all Recall levels, as shown in Equation (\ref{eq:AP}):
\begin{equation}
\label{eq:AP}
\text{AP} = \int_{0}^{1} \text{Precision}(\text{Recall})\,d(\text{Recall})
\end{equation}

The Mean Average Precision (mAP) is the average of the AP values across all N object categories, providing a holistic measure of detection accuracy across all classes, as shown in Equation (\ref{eq:mAP}):
\begin{equation}
\label{eq:mAP}
\text{mAP} = \frac{1}{N} \sum_{i=1}^{N} \text{AP}_{i}
\end{equation}
where the $AP_{i}$ in the equation denotes the AP value with category index value $i$, and N denotes the number of categories of the samples in the training dataset, in this paper, N is 8 and 12 for HiXray and PIDray respectively. Specifically, mAP50 denotes the mAP when the IoU threshold is fixed at 0.50, serving as a robust baseline for overall performance. mAP50:95 is a more stringent metric, representing the average mAP across IoU thresholds ranging from 0.50 to 0.95 with 0.05 increments. This multi-threshold averaging allows mAP50:95 to comprehensively reflect the model's ability to accurately localize objects under varying degrees of overlap and strict bounding box requirements.

For model efficiency, we use GFLOPs to measure the theoretical computational complexity of the model, as shown in Equation (\ref{eq:GFLOPs}). GFLOPs quantifies the total number of floating-point operations required to process a single image, providing an insight into the inherent efficiency of the model architecture. Specifically, GFLOPs is calculated as the total number of floating-point operations (FLOPs) divided by \(10^9\), which standardizes the value to a more manageable scale. Moreover, Frames Per Second (FPS) measures the actual inference speed on the target hardware. It is the reciprocal of the average inference time \(\text{T}_{\text{inference}}\) per image, as shown in Equation (\ref{eq:FPS}), and measures the model's real-time processing capability in practical security inspection scenarios:
\begin{equation}
\label{eq:GFLOPs}
\text{GFLOPs}=\frac{\text{Floating Point per Operations}}{10^{9}}
\end{equation}

\begin{equation}
\label{eq:FPS}
\text{FPS}=\frac{1}{\text{T}_{\text{inference}}}
\end{equation}

\subsubsection{Implementation Details}
The GSA-YOLO framework is built upon the lightweight YOLOv8n architecture, while YOLOv8m was selected as the teacher model for the Ada-KD mechanism as discussed in Section \ref{subsec:comparison experiment}. Our proposed GSA-YOLO and the YOLOv8n baseline were implemented via the Ultralytics Python package (version 8.3.199) \cite{ultralytics_yolov8}. All experiments, including the training and evaluation of comparative models reported in Tables \ref{tab:hixray_results} and \ref{tab:pidray_results}, were executed on a consistent computational platform anchored by an NVIDIA RTX 3090 GPU, utilizing Python 3.10.18 and PyTorch 2.7.1 integrated with CUDA 12.6. For model training, the input image size was fixed at 640 × 640, a batch size of 64 was employed, and the YOLOv8n baseline was trained for 100 epochs on both datasets. \revise{Our multi-stage training pipeline consists of three sequential phases: (1) Group Lasso training using SGD optimizer (lr=1e-4, momentum=0.9, weight decay=5e-4) with cosine annealing scheduler for 60 epochs, with the Backbone weights remaining unfrozen and jointly optimized throughout this stage; (2) SSS training initialized from the GL-trained model weights, using the same SGD configuration with gamma schedule (10 epochs warmup, 40 epochs ramp to target gamma=1e-4) for 85 epochs, where pruning is performed at the end of sparse training and no separate fine-tuning stage is added after pruning since the Ada-KD stage serves this recovery function; (3) Ada-KD training initialized from the SSS-pruned model, using SGD optimizer with cosine annealing scheduler for 60 epochs, totaling 205 epochs across all stages. Notably, during the SSS training stage, the GL-trained Neck weights are not frozen but continue to be jointly optimized, ensuring that the feature distribution of the Neck remains adaptive and does not adversely affect the evaluation of channel importance in the Head.} To ensure the reliability and fair comparison of efficiency metrics, all models were tested using the same 640 × 640 input resolution at a batch size of 1, running in standard FP32 precision. A warm-up phase of 50 forward passes was executed prior to formal timing to stabilize GPU performance. \revise{The formal FPS measurement is then conducted over the full test set (9096 images for HiXray and 18220 images for PIDray) with continuous sequential inference, and the reported FPS is computed as the average frame rate over all test images.} The reported FPS specifically reflects the latency of the model's forward pass, the inference time, without including post-processing operations like NMS or box decoding. The detailed experiment configuration and settings are shown in Table \ref{tab:exp_config}.

\begin{table}[htbp]
\centering
\caption{Detailed experimental configuration and settings}\label{tab:exp_config}
\resizebox{\columnwidth}{!}{ 
\begin{tabular}{ll} 
\toprule
Configuration & Setting \\
\midrule
Operating System & Ubuntu 20.04 \\
CPU & Intel(R) Core(TM) Ultra 7 255H 2.00 GHz \\
GPU & NVIDIA RTX 3090 \\
CUDA & 12.6 \\
Framework & PyTorch 2.7.1 \\
Language & Python 3.10.18 \\
Input Resolution (Training \& Test) & $640 \times 640$ \\
Training Batch Size & 64 \\
Inference Batch Size & 1 \\
Measurement Precision & FP32 (32-bit floating-point) \\
FPS Measurement Scope & Forward Pass Latency Only \\
GPU Pre-Heating & 50 Warm-up Passes \\
\bottomrule
\end{tabular}
} 
\end{table}

\subsection{Comparison Experiment}
\label{subsec:comparison experiment}
The effectiveness of the proposed GSA-YOLO was rigorously validated by comparing it with mainstream methods for detecting prohibited items and several well-known object detection models \cite{noauthor_fine-yolo_nodate,noauthor_yolo-srw_nodate,haq_esi-yolo_2025,zhang_dsf-yolo_2025,wang_improved_2024}on the HiXray and PIDray datasets. \revise{Our comparison strategy focuses on X-ray security detection methods and established baselines to ensure domain-relevant evaluation.} \revise{All reported results represent averages from multiple runs with different random seeds to ensure statistical reliability.}  The evaluation metrics included GFLOPs, FPS, Precision, Recall, mAP50, and mAP50:95, covering both computational efficiency and detection accuracy.  As presented in Table \ref{tab:hixray_results} and Table \ref{tab:pidray_results}, GSA-YOLO consistently achieves a new competitive balance between efficiency and accuracy. \revise{The proposed method reduces computational cost from 8.7G to 8.0G GFLOPs while maintaining competitive performance, achieving an effective efficiency-accuracy trade-off.}
\begin{table*}[htbp]
\centering
\caption{Evaluation results of the different methods on the HiXray dataset. (The best performance in each metric is marked in bold, and the second-best is underlined.)}\label{tab:hixray_results}
\resizebox{\textwidth}{!}{%
\begin{tabular}{lccccccc} 
\toprule
Model & Model Size & GFLOPs & FPS & Precision & Recall & mAP50 & mAP50:95 \\
\midrule
Faster R-CNN\cite{ren_faster_2015} & 319.2MB & 168.7G & 15.02 & 0.841 & 0.763 & 0.788 & 0.467 \\
RetinaNet\cite{lin_focal_2017} & 128.7MB & 151.5G & 34.05 & 0.824 & 0.723 & 0.756 & 0.451 \\
YOLOv5n\cite{jocher_ultralyticsyolov5_2020} & \textbf{3.9MB} & \textbf{4.5G} & 161.31 & 0.859 & 0.773 & 0.799 & 0.486 \\
YOLOv8n\cite{ultralytics_yolov8} & 6.2MB & 8.7G & \underline{170.58} & 0.882 & 0.769 & 0.806 & 0.507 \\
\newrevise{YOLOv8n-Pruned (L1)} & \newrevise{4.4MB} & \newrevise{6.1G} & \newrevise{231.92} & \newrevise{0.869} & \newrevise{0.760} & \newrevise{0.795} & \newrevise{0.496} \\
\newrevise{YOLOv8n-Pruned+QAT (INT8)} & \newrevise{2.3MB} & \newrevise{6.1G} & \newrevise{312.50} & \newrevise{0.856} & \newrevise{0.748} & \newrevise{0.782} & \newrevise{0.487} \\
\revise{YOLOv8m}\cite{ultralytics_yolov8} & \revise{50.3MB} & \revise{79.1G} & \revise{96.23} & \revise{0.912} & \revise{0.806} & \revise{0.839} & \revise{0.550} \\
Fine-YOLO\cite{noauthor_fine-yolo_nodate} & 46.2MB & 56.9G & 152.42 & 0.877 & 0.771 & 0.811 & 0.514 \\
SRW-YOLO\cite{noauthor_yolo-srw_nodate} & 9.4MB & 12.0G & 158.97 & 0.883 & 0.782 & 0.814 & 0.516 \\
ESI-YOLO\cite{haq_esi-yolo_2025} & 6.7MB & 8.9G & 164.48 & 0.886 & 0.779 & 0.815 & 0.517 \\
DSF-YOLO\cite{zhang_dsf-yolo_2025} & 35.4MB & 43.4G & 54.60 & 0.889 & 0.786 & 0.817 & 0.515 \\
GEMA-YOLO\cite{wang_improved_2024} & 11.8MB & 16.1G & 82.57 & \underline{0.895} & \underline{0.788} & \underline{0.822} & \underline{0.523} \\
GSA-YOLO(ours) & \underline{5.7MB} & \underline{8.0G} & \textbf{189.62} & \textbf{0.902} & \textbf{0.795} & \textbf{0.827} & \textbf{0.531} \\
\bottomrule
\end{tabular}%
}
\end{table*}

\begin{table*}[htbp] 
\centering
\caption{Evaluation results of the different methods on the PIDray dataset. (The best performance in each metric is marked in bold, and the second-best is underlined.)}\label{tab:pidray_results}
\resizebox{\textwidth}{!}{%
\begin{tabular}{lccccccccc} 
\toprule
\multirow{2}{*}{Model} & \multirow{2}{*}{Model Size} & \multicolumn{2}{c}{easy} & \multicolumn{2}{c}{hard} & \multicolumn{2}{c}{hidden} & \multicolumn{2}{c}{Average} \\ 
\cline{3-10}
& & mAP50 & mAP50:95 & mAP50 & mAP50:95 & mAP50 & mAP50:95 & mAP50 & mAP50:95 \\
\midrule
Faster R-CNN\cite{ren_faster_2015} & 319.2MB & 0.805 & 0.679 & 0.820 & 0.632 & 0.582 & 0.403 & 0.736 & 0.571 \\
Retina-Net\cite{lin_focal_2017} & 128.7MB & 0.799 & 0.672 & 0.813 & 0.621 & 0.575 & 0.401 & 0.729 & 0.565 \\
YOLOv5n\cite{jocher_ultralyticsyolov5_2020} & \textbf{3.9MB} & 0.817 & 0.691 & 0.847 & 0.651 & 0.593 & 0.406 & 0.752 & 0.583 \\
YOLOv8n\cite{ultralytics_yolov8} & 6.2MB & 0.844 & 0.750 & 0.862 & 0.718 & 0.644 & 0.514 & 0.783 & 0.661 \\
\newrevise{YOLOv8n-Pruned (L1)} & \newrevise{4.4MB} & \newrevise{0.832} & \newrevise{0.741} & \newrevise{0.851} & \newrevise{0.705} & \newrevise{0.635} & \newrevise{0.504} & \newrevise{0.769} & \newrevise{0.648} \\
\newrevise{YOLOv8n-Pruned+QAT (INT8)} & \newrevise{2.3MB} & \newrevise{0.822} & \newrevise{0.729} & \newrevise{0.838} & \newrevise{0.696} & \newrevise{0.622} & \newrevise{0.492} & \newrevise{0.758} & \newrevise{0.638} \\
\revise{YOLOv8m}\cite{ultralytics_yolov8} & \revise{50.3MB} & \revise{0.876} & \revise{0.781} & \revise{0.888} & \revise{0.745} & \revise{0.674} & \revise{0.533} & \revise{0.813} & \revise{0.675} \\
Fine-YOLO\cite{noauthor_fine-yolo_nodate} & 46.2MB & 0.848 & 0.752 & 0.866 & 0.721 & 0.646 & 0.512 & 0.787 & 0.662 \\
SRW-YOLO\cite{noauthor_yolo-srw_nodate} & 9.4MB & 0.847 & 0.754 & 0.868 & 0.722 & 0.642 & 0.509 & 0.786 & 0.662 \\
ESI-YOLO\cite{haq_esi-yolo_2025} & 6.7MB & 0.850 & 0.759 & 0.871 & 0.724 & 0.651 & \underline{0.518} & 0.791 & 0.667 \\
DSF-YOLO\cite{zhang_dsf-yolo_2025} & 35.4MB & 0.852 & 0.756 & \underline{0.876} & 0.724 & 0.648 & 0.516 & 0.792 & 0.665 \\
GEMA-YOLO\cite{wang_improved_2024} & 11.8MB & \underline{0.861} & \underline{0.763} & 0.875 & \underline{0.729} & \underline{0.654} & 0.517 & \underline{0.797} & \underline{0.670} \\
GSA-YOLO(ours) & \underline{5.7MB} & \textbf{0.868} & \textbf{0.771} & \textbf{0.879} & \textbf{0.734} & \textbf{0.665} & \textbf{0.524} & \textbf{0.804} & \textbf{0.679} \\
\bottomrule
\end{tabular}
} 
\end{table*}

On the HiXray dataset, GSA-YOLO achieves competitive performance across the evaluated metrics. The inference speed reaches 189.62 FPS, \revise{an 11\% improvement over YOLOv8n, attributed to the SSS module which reduces GFLOPs from 8.7G to 8.0G through structured pruning of redundant channels in the detection head}. In terms of detection accuracy, GSA-YOLO achieves \revise{mAP50 of 0.827 and mAP50:95 of 0.531, representing improvements of 2.1\% and 2.4\% over the YOLOv8n baseline}. \revise{The Ada-KD mechanism contributes to accuracy recovery by transferring soft-label knowledge from the teacher model, mitigating the degradation introduced during the sparsity stages}.

On the PIDray dataset, which includes easy, hard, and hidden test partitions, GSA-YOLO achieves the best performance across all eight evaluated metrics. \revise{On the hidden test set, the model obtains mAP50 of 0.665 and mAP50:95 of 0.524. The GL module, by applying soft regularization to the neck's feature fusion layers, helps preserve multi-scale feature representations under occlusion and cluttered backgrounds}. Overall, GSA-YOLO achieves \revise{mAP50 of 0.804 and mAP50:95 of 0.679, exceeding the baseline by 2.1\% and 1.8\% respectively}. \newrevise{Compared to the two compression baselines, GSA-YOLO outperforms YOLOv8n-Pruned by +0.037 mAP50:95 on HiXray and +0.032 on PIDray, and exceeds YOLOv8n-Pruned+QAT by +0.050 and +0.045 respectively, confirming that Group Lasso-guided sparsity with Ada-KD recovery achieves a superior accuracy-efficiency trade-off over aggressive compression without knowledge distillation.}

\revise{To select an appropriate teacher model for Ada-KD, we compare the YOLOv8 series from YOLOv8n to YOLOv8l in terms of the trade-off between distillation gain and training overhead.}

\begin{figure}[ht]
    \centering
    \includegraphics[width=0.85\linewidth, keepaspectratio]{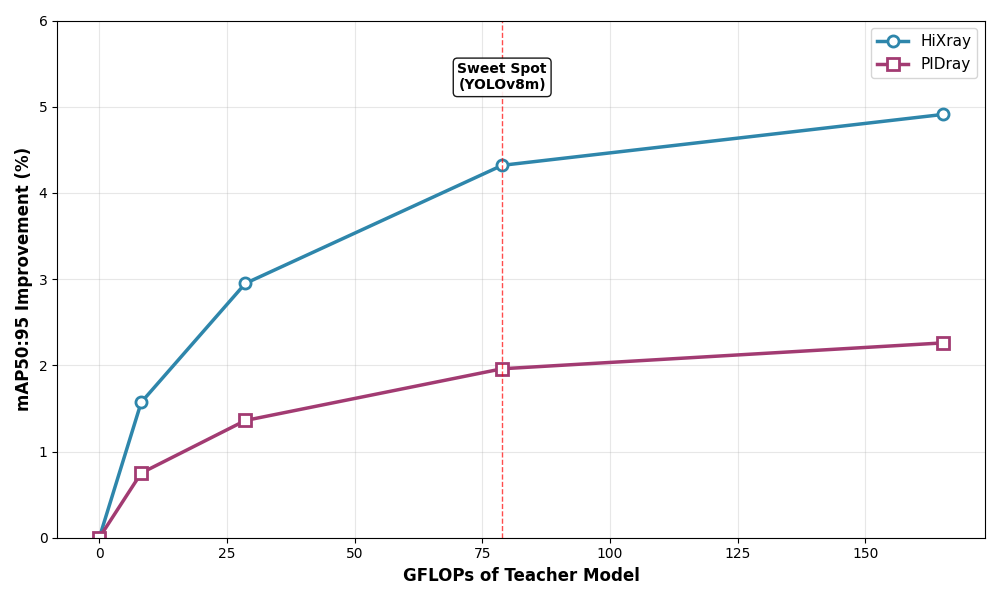}
    \caption{The trade-off analysis of teacher model complexity (GFLOPs) vs. student model performance (mAP50:95) for Ada-KD mechanism}
    \label{fig:teacher_student}
\end{figure}

As illustrated in Figure \ref{fig:teacher_student}, \revise{the student model's mAP50:95 plateaus beyond YOLOv8m, while training time continues to increase, making YOLOv8m a reasonable choice that balances accuracy recovery and training cost}. We thus selected YOLOv8m as our teacher model in our Ada-KD mechanism.

\subsection{Hyperparameter Sensitivity Analysis}
\label{subsec:hyperparameter sensitivity analysis}
\begin{figure}[ht]
    \centering
    \subfloat[Sensitivity to GL parameter $\beta$.\label{fig:beta_analysis}]{\includegraphics[width=0.48\textwidth]{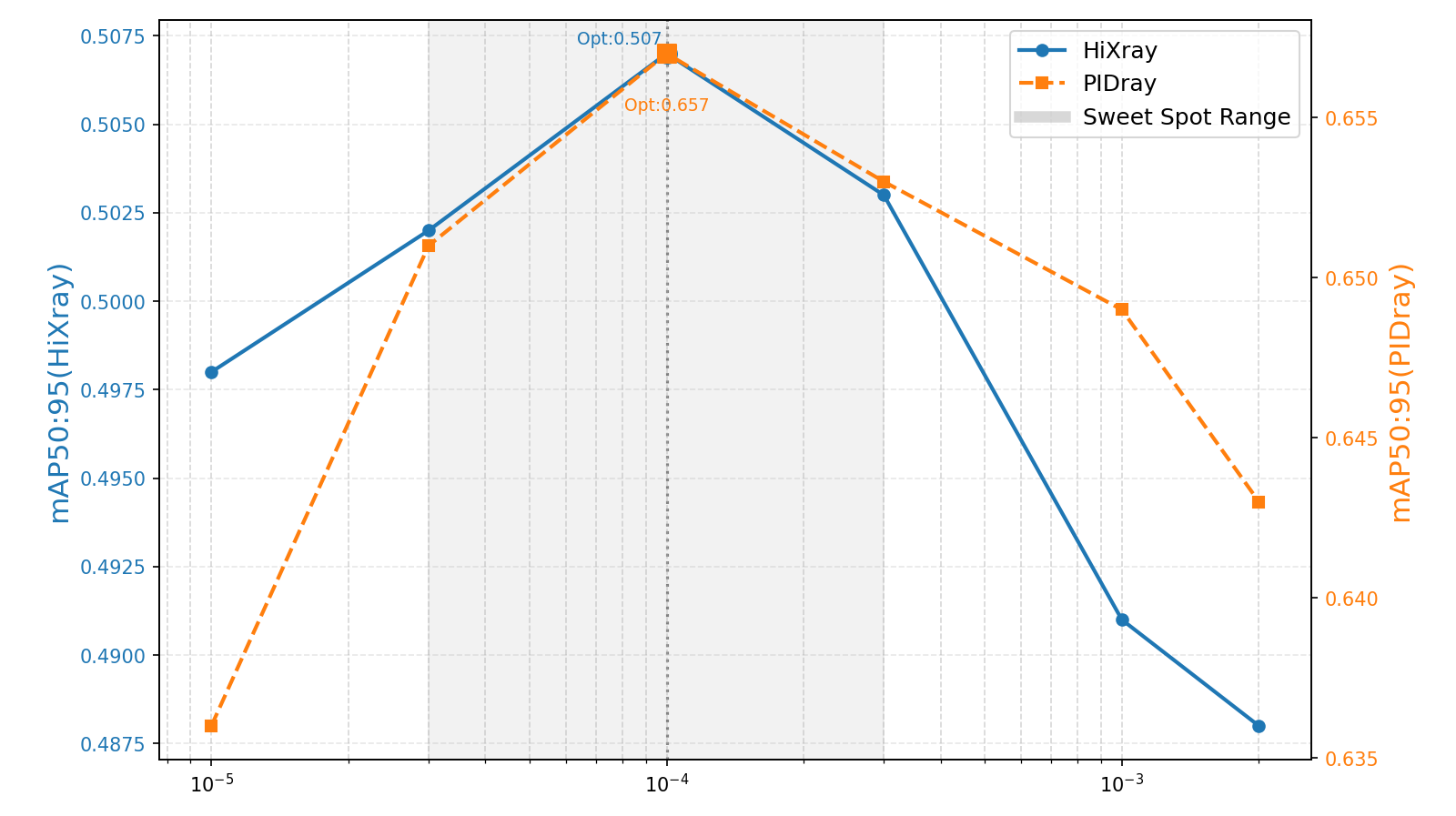}} 
    \hspace{0.01\textwidth} 
    \subfloat[Sensitivity to SSS parameter $\gamma$.\label{fig:gamma_analysis}]{\includegraphics[width=0.48\textwidth]{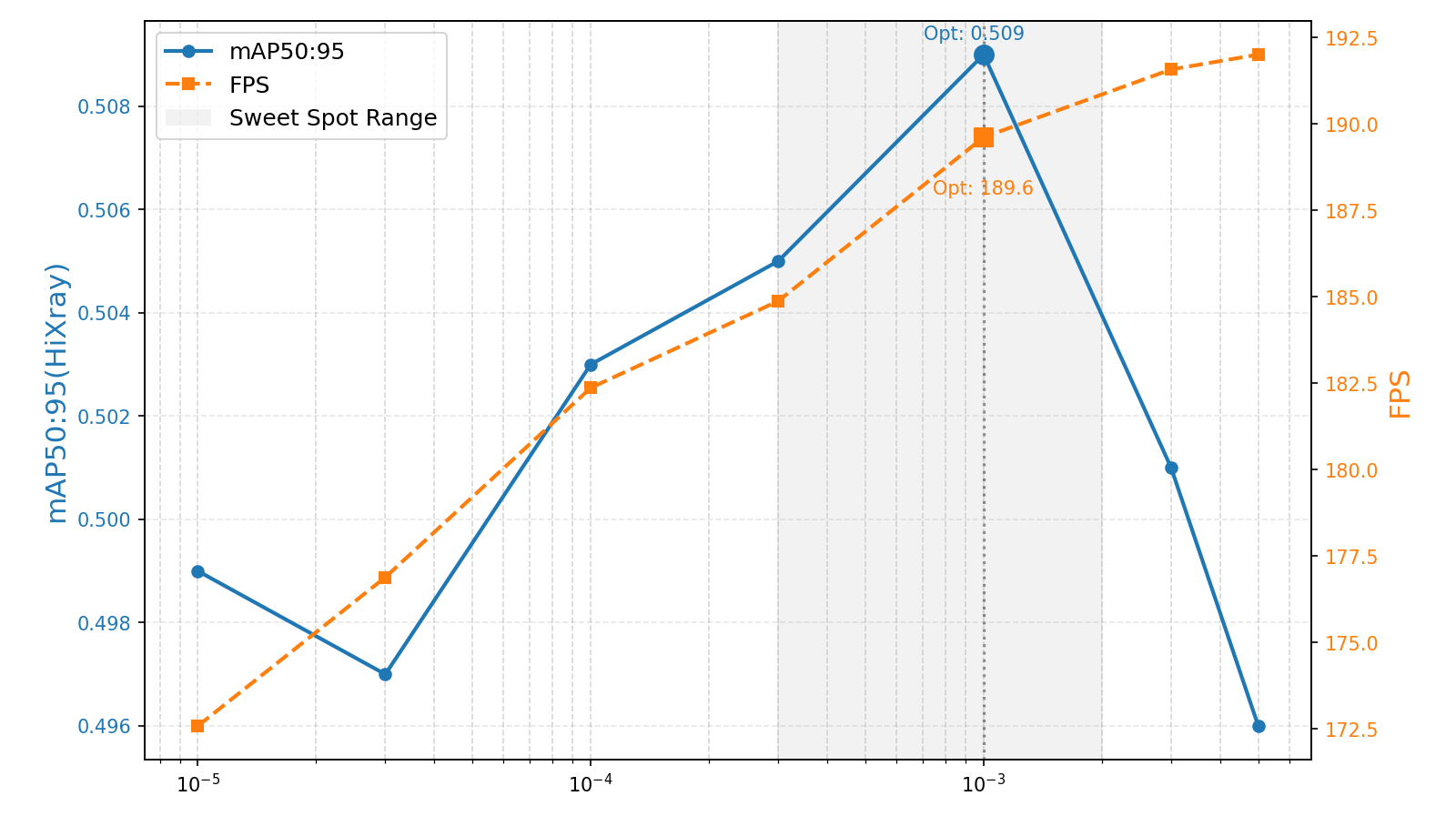}} \\
    
    \vspace{0.3cm} 
    
    \subfloat[Sensitivity to Ada-KD parameter $\lambda_0$.\label{fig:lambda0_analysis}]{\includegraphics[width=0.48\textwidth]{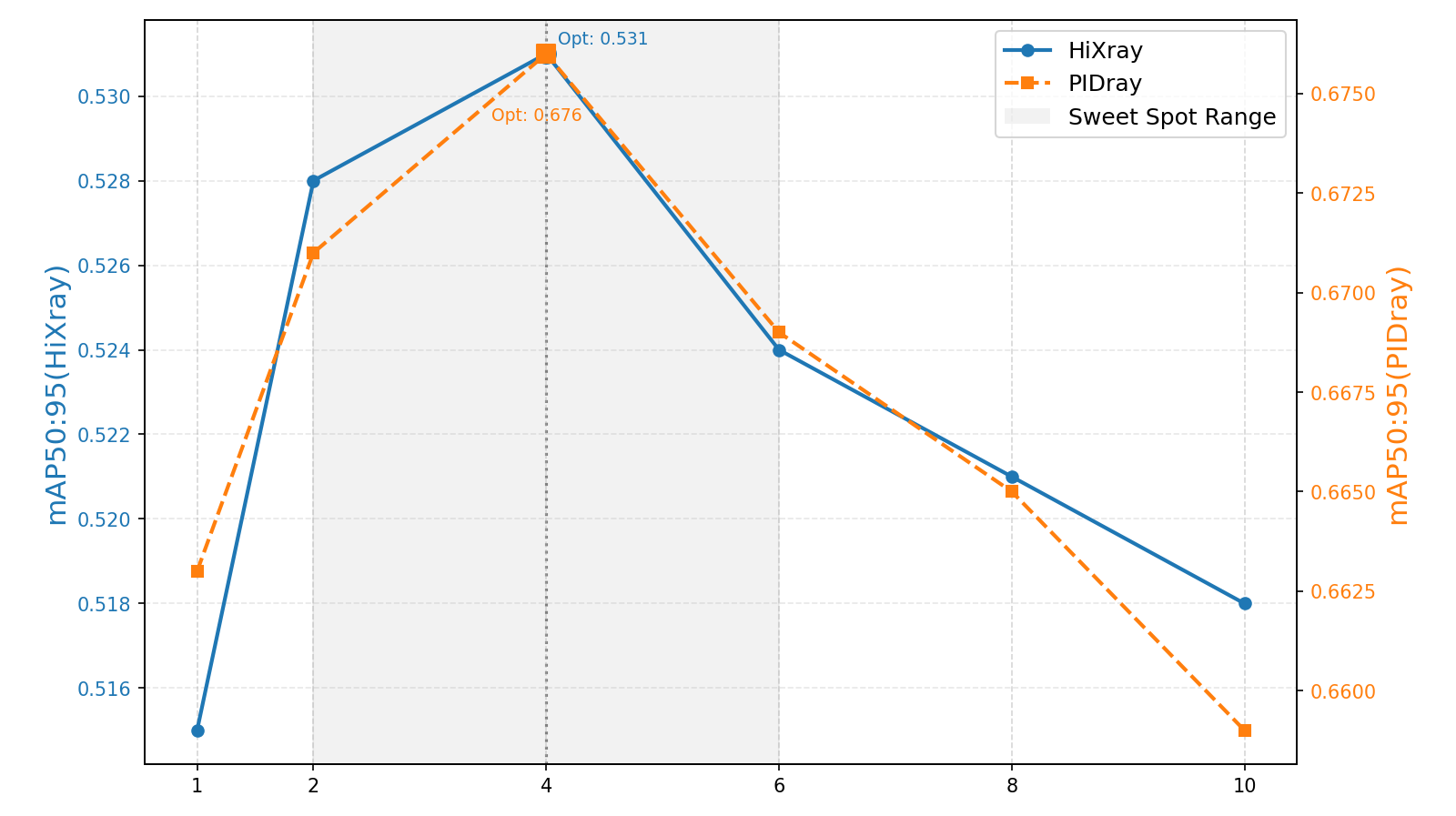}} 
    \hspace{0.01\textwidth} 
    \subfloat[Sensitivity to Ada-KD parameter $\theta$.\label{fig:theta_analysis}]{\includegraphics[width=0.48\textwidth]{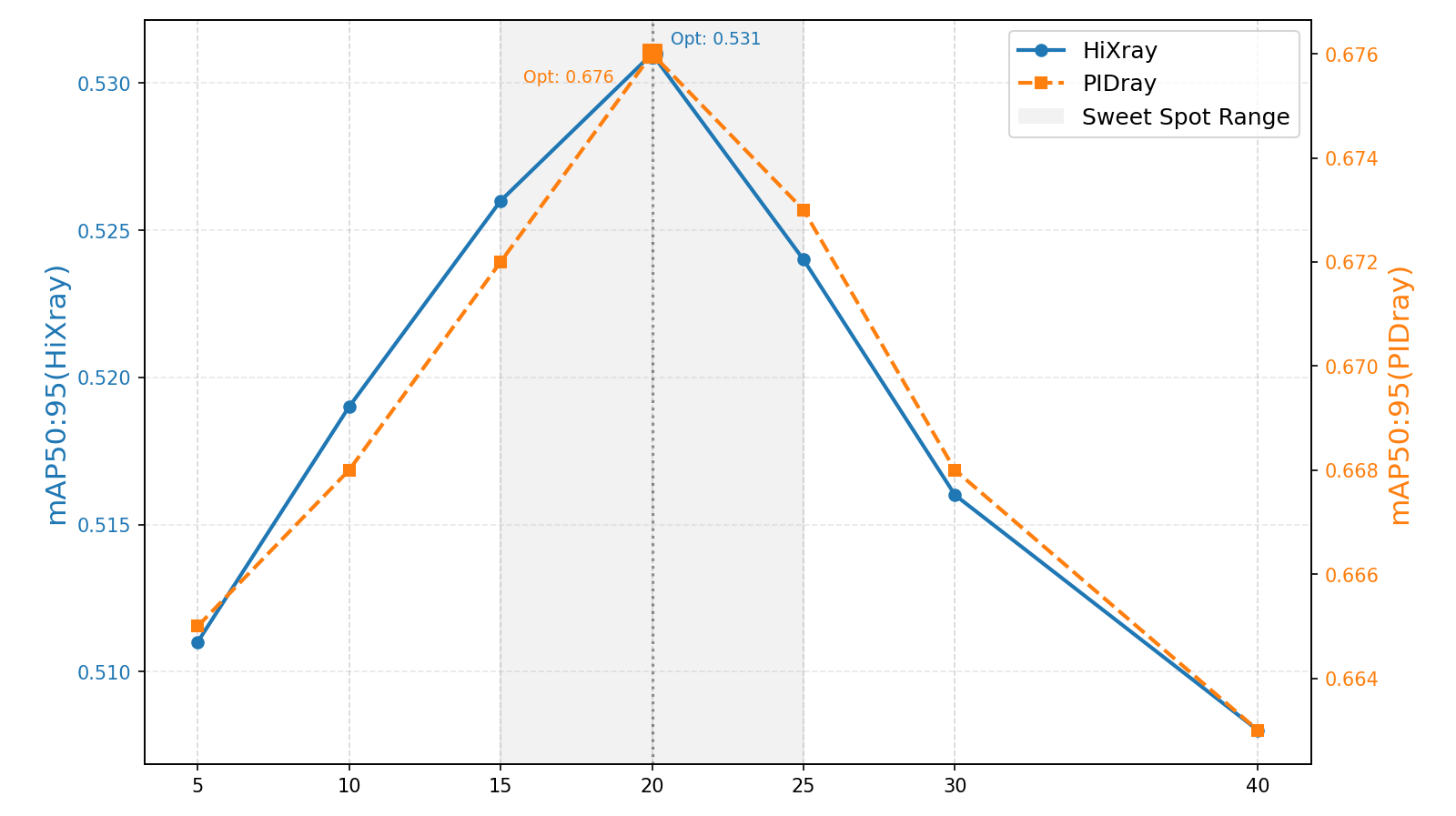}}
    
    \caption{\revise{Hyperparameter sensitivity analysis for the GSA-YOLO framework. The shaded regions in each subplot represent the optimal parameter ranges where the models achieve balance between performance and efficiency}.}
    \label{fig:sensitivity_analysis}
\end{figure}

\revise{To optimize the GSA-YOLO framework performance and guide practical deployment, we conducted hyperparameter sensitivity analysis across the three core modules: GL, SSS and Ada-KD}. \revise{The objective is to identify optimal parameter values that balance detection accuracy and efficiency}. \revise{The analysis follows rigorous experimental protocols: the control variable method is adopted so that only one hyperparameter is varied at a time while all others remain fixed at their baseline values; a fixed random seed is used throughout all sensitivity experiments to ensure reproducibility; grid search is employed over predefined value ranges for each hyperparameter; and all configurations are evaluated on both HiXray and PIDray datasets under identical training procedures.} The tested ranges and optimal values for all hyperparameters are summarized in Table \ref{tab:hyperparameters}. The detailed trends and trade-offs are visualized in Figure \ref{fig:sensitivity_analysis}.

\begin{table}[ht]
    \centering
    \caption{Summary of hyperparameter sensitivity analysis}
    \label{tab:hyperparameters}
    \begin{tabular}{ccccc}
        \toprule
        \textbf{Module} & \textbf{Hyperparameter} &  \textbf{Search Range} & \textbf{Optimal Value} \\
        \midrule
        GL & $\beta$ & $[1\mathrm{e}{-5}, 2\mathrm{e}{-3}]$ & $1\mathrm{e}{-4}$ \\
        SSS & $\gamma$ &  $[1\mathrm{e}{-5}, 5\mathrm{e}{-3}]$ & $1\mathrm{e}{-3}$ \\
        Ada-KD & $\lambda_0$ & $\{1, 2, 4, 6, 8, 10\}$ & $4$ \\
        Ada-KD & $\theta$ & $\{5, 10, 15, 20, 25, 30, 40\}$ & $15$ \\
        \bottomrule
    \end{tabular}
\end{table}

\revise{The GL parameter controls soft sparsity regularization intensity for feature refinement in the neck part. Performance curves on both HiXray and PIDray show a convex shape, with peak performance at $\beta=1\mathrm{e}{-4}$. This value provides regularization strength that maintains model performance against complex clutter while preserving feature pathways}. As shown in Figure \ref{fig:beta_analysis}, \revise{the optimal performance range corresponds to $[3\mathrm{e}{-5}, 5\mathrm{e}{-4}]$}.

\revise{For the SSS module, we examine the pruning intensity $\gamma$ as shown in Figure} \ref{fig:gamma_analysis}\revise{, which affects the speed-accuracy relationship. The plot uses dual Y-axes to track mAP50:95 (left) and FPS (right). As $\gamma$ increases, FPS shows continuous improvement due to model slimming. mAP50:95 peaks at $\gamma=1\mathrm{e}{-3}$, followed by decline, indicating that pruning beyond $\gamma > 3\mathrm{e}{-3}$ reduces model capacity. The value $\gamma=1\mathrm{e}{-3}$ is selected for balancing FPS improvement and accuracy maintenance}.

\revise{In addition to $\gamma$, the pruning threshold $\tau$ directly determines which channels are removed after SSS sparse training.} Table~\ref{tab:tau_sensitivity} \revise{presents the sensitivity analysis of $\tau$ on both datasets. The results show that $\tau = 0.001$ consistently achieves the best performance, while smaller values provide insufficient pruning and larger values cause over-aggressive channel removal with measurable accuracy degradation.}

\begin{table}[htbp]
\centering
\caption{\revise{Sensitivity analysis of pruning threshold $\tau$ on HiXray and PIDray datasets.}}\label{tab:tau_sensitivity}
\resizebox{\textwidth}{!}{
\begin{tabular}{c|cccc|cccc}
\hline
\multirow{2}{*}{\revise{$\tau$}} & \multicolumn{4}{c|}{\revise{HiXray}} & \multicolumn{4}{c}{\revise{PIDray}} \\
 & \revise{Precision} & \revise{Recall} & \revise{mAP50} & \revise{mAP50:95} & \revise{Precision} & \revise{Recall} & \revise{mAP50} & \revise{mAP50:95} \\
\hline
\revise{0.0005} & \revise{0.897} & \revise{0.791} & \revise{0.822} & \revise{0.526} & \revise{0.832} & \revise{0.736} & \revise{0.799} & \revise{0.674} \\
\revise{\textbf{0.0010}} & \revise{\textbf{0.902}} & \revise{\textbf{0.795}} & \revise{\textbf{0.827}} & \revise{\textbf{0.531}} & \revise{\textbf{0.837}} & \revise{\textbf{0.741}} & \revise{\textbf{0.804}} & \revise{\textbf{0.679}} \\
\revise{0.0015} & \revise{0.899} & \revise{0.792} & \revise{0.824} & \revise{0.528} & \revise{0.834} & \revise{0.738} & \revise{0.801} & \revise{0.676} \\
\revise{0.0020} & \revise{0.894} & \revise{0.787} & \revise{0.820} & \revise{0.523} & \revise{0.830} & \revise{0.733} & \revise{0.797} & \revise{0.670} \\
\revise{0.0030} & \revise{0.886} & \revise{0.780} & \revise{0.813} & \revise{0.515} & \revise{0.821} & \revise{0.726} & \revise{0.789} & \revise{0.662} \\
\hline
\end{tabular}
}
\end{table}

\revise{The Ada-KD module involves two interdependent parameters: Maximum Distillation Intensity $\lambda_0$ and Decay Steepness $\theta$. Performance for both datasets peaks at $\lambda_0=4$, which falls within the range shown in Figure} \ref{fig:lambda0_analysis}. \revise{Low $\lambda_0$ values limit knowledge transfer, while high values cause over-distillation, affecting the student model's learning of hard targets. For $\theta$, which controls the Ada-KD weight decay rate during training, results show that $\theta=15$ yields the highest mAP50:95. This value balances knowledge transfer in early phases with student model optimization during convergence. The performance range corresponds to $[15,25]$ as illustrated in Figure} \ref{fig:theta_analysis}.

\revise{We further conduct sensitivity analysis on the inflection point $T_{\text{mid}}$ of the adaptive weight function. As shown in Table}~\ref{tab:tmid_sensitivity}\revise{, $T_{\text{mid}}=0.5$ consistently achieves the best performance on HiXray, validating that an equal split between teacher-guided and student-independent phases provides the optimal knowledge transfer schedule for X-ray security inspection scenarios.}

\begin{table}[htbp]
\centering
\caption{\revise{Sensitivity analysis of inflection point $T_{\text{mid}}$ on HiXray dataset.}}\label{tab:tmid_sensitivity}
\resizebox{0.75\textwidth}{!}{
\begin{tabular}{c|cccc}
\hline
\revise{$T_{\text{mid}}$} & \revise{Precision} & \revise{Recall} & \revise{mAP50} & \revise{mAP50:95} \\
\hline
\revise{0.3} & \revise{0.896} & \revise{0.788} & \revise{0.820} & \revise{0.525} \\
\revise{0.4} & \revise{0.900} & \revise{0.792} & \revise{0.824} & \revise{0.529} \\
\revise{\textbf{0.5}} & \revise{\textbf{0.902}} & \revise{\textbf{0.795}} & \revise{\textbf{0.827}} & \revise{\textbf{0.531}} \\
\revise{0.6} & \revise{0.899} & \revise{0.793} & \revise{0.825} & \revise{0.528} \\
\revise{0.7} & \revise{0.895} & \revise{0.788} & \revise{0.820} & \revise{0.523} \\
\hline
\end{tabular}
}
\end{table}

\revise{To validate the decay-type weight design, we compare it against alternative weighting patterns on the HiXray dataset. As shown in Table}~\ref{tab:weight_pattern}\revise{, the decay pattern consistently outperforms increasing, bell-shaped, and fixed alternatives, confirming that strong initial teacher guidance followed by gradual student independence is the most effective strategy for accuracy recovery in sparse X-ray detection models.}

\begin{table}[htbp]
\centering
\caption{\revise{Comparison of Ada-KD weight design patterns on HiXray dataset.}}\label{tab:weight_pattern}
\resizebox{\textwidth}{!}{
\begin{tabular}{l|l|cccc}
\hline
\revise{Pattern} & \revise{Description} & \revise{Precision} & \revise{Recall} & \revise{mAP50} & \revise{mAP50:95} \\
\hline
\revise{\textbf{Decay (Ours)}} & \revise{High$\to$Low (0.8$\to$0.2)} & \revise{\textbf{0.902}} & \revise{\textbf{0.795}} & \revise{\textbf{0.827}} & \revise{\textbf{0.531}} \\
\revise{Increasing} & \revise{Low$\to$High (0.2$\to$0.8)} & \revise{0.892} & \revise{0.783} & \revise{0.816} & \revise{0.521} \\
\revise{Bell-shaped} & \revise{Low$\to$High$\to$Low (0.2$\to$0.8$\to$0.2)} & \revise{0.896} & \revise{0.787} & \revise{0.820} & \revise{0.524} \\
\revise{Fixed} & \revise{Constant (0.5)} & \revise{0.888} & \revise{0.779} & \revise{0.813} & \revise{0.517} \\
\hline
\end{tabular}
}
\end{table}

\subsection{Ablation Experiment}
\label{subsec:ablation experiment}
\revise{Ablation experiments were conducted on both datasets to evaluate the individual and combined impact of the modules: GL, SSS and Ada-KD on YOLOv8n baseline performance}. Precision, Recall, mAP50 and mAP50:95 were selected as evaluation metrics. The composition and description of the module groups utilized in the experiments are as follows:
\begin{itemize}
    \item[$\circ$] \textbf{Baseline:} The original YOLOv8n model, serving as the benchmark without any proposed modifications.
    \item[$\circ$] \textbf{GL:} Implementation of the GL module solely on the Neck of the YOLOv8n model, designed for initial feature refinement and soft sparsity.
    \item[$\circ$] \textbf{SSS:} Implementation of the SSS module solely on the detection head of the YOLOv8n model, \revise{applying hard pruning for efficiency}.
    \item[$\circ$] \textbf{GL+SSS:} The combination of GL on the Neck and SSS on the Head, \revise{creating a structurally compact and sparse architecture}.
    \item[$\circ$] \textbf{GL+Ada-KD:} Applying the Ada-KD mechanism on the GL-integrated model, \revise{targeting recovery of neck-refined features}.
    \item[$\circ$] \textbf{SSS+Ada-KD:} Applying the Ada-KD mechanism on the SSS-integrated model, \revise{targeting recovery of head-pruned channels}.
    \item[$\circ$] \textbf{GL+SSS+Ada-KD (GSA-YOLO):} \revise{The complete model integrating all three components for efficiency, accuracy and robustness}.
\end{itemize}

\begin{table}[ht]
    \centering
\caption{Ablation experiment on HiXray dataset}\label{tab:ablation_hixray}
\begin{tabular}{ccccccc} 
\toprule
GL & SSS & Ada-KD & Precision & Recall & mAP50 & mAP50:95 \\
\midrule
$\times$ & $\times$ & $\times$ & 0.882 & 0.769 & 0.806 & 0.507 \\
$\checkmark$ & $\times$ & $\times$ & 0.871 & 0.768 & 0.809 & 0.501 \\
$\times$ & $\checkmark$ & $\times$ & 0.873 & 0.776 & 0.808 & 0.504 \\
$\checkmark$ & $\checkmark$ & $\times$ & 0.885 & 0.781 & 0.811 & 0.509 \\
$\checkmark$ & $\times$ & $\checkmark$ & 0.886 & 0.786 & 0.813 & 0.513 \\
$\times$ & $\checkmark$ & $\checkmark$ & 0.888 & 0.789 & 0.815 & 0.514 \\
$\checkmark$ & $\checkmark$ & $\checkmark$ & \textbf{0.902} & \textbf{0.795} & \textbf{0.827} & \textbf{0.531} \\
\bottomrule
\end{tabular}
\end{table}

\begin{table}[ht]
\centering
\caption{Ablation experiment on PIDray dataset}\label{tab:ablation_pidray}
\begin{tabular}{ccccccc} 
\toprule
GL & SSS & Ada-KD & Precision & Recall & mAP50 & mAP50:95 \\
\midrule
$\times$ & $\times$ & $\times$ & 0.825 & 0.724 & 0.783 & 0.661 \\
$\checkmark$ & $\times$ & $\times$ & 0.820 & 0.728 & 0.781 & 0.657 \\
$\times$ & $\checkmark$ & $\times$ & 0.818 & 0.721 & 0.777 & 0.652 \\
$\checkmark$ & $\checkmark$ & $\times$ & 0.830 & 0.731 & 0.788 & 0.663 \\
$\checkmark$ & $\times$ & $\checkmark$ & 0.828 & 0.734 & 0.789 & 0.665 \\
$\times$ & $\checkmark$ & $\checkmark$ & 0.827 & 0.729 & 0.785 & 0.664 \\
$\checkmark$ & $\checkmark$ & $\checkmark$ & \textbf{0.837} & \textbf{0.741} & \textbf{0.804} & \textbf{0.679} \\
\bottomrule
\end{tabular}
\end{table}

Table \ref{tab:ablation_hixray} presents the ablation results on the HiXray dataset. Applied individually, GL and SSS each show only marginal differences from the Baseline: SSS slightly improves Recall to $0.776$, while GL achieves an mAP50 of $0.809$, but both also introduce a minor accuracy trade-off during the compression phase. The combination of GL+SSS shows complementary effects, surpassing the Baseline across all metrics with mAP50 of $0.811$ and mAP50:95 of $0.509$.

Ada-KD consistently mitigates the accuracy degradation introduced during compression. GL+Ada-KD improves upon the GL-only model by 1.4\% in Recall and 1.2\% in mAP50:95, while SSS+Ada-KD achieves a 1.5\% and 1.3\% increase in Precision and Recall over SSS-only, confirming Ada-KD's role in recovering performance in compressed models. The complete GL+SSS+Ada-KD (GSA-YOLO) architecture achieves the best overall performance, with gains of 2.0\% in Precision, 2.6\% in Recall, 2.1\% in mAP50, and 2.4\% in mAP50:95 over the Baseline. \revise{These improvements represent relative gains of 4.7\% on HiXray and 2.7\% on PIDray in terms of mAP50:95.}

Table \ref{tab:ablation_pidray} presents the ablation results on the PIDray dataset. The trend is consistent with HiXray: individual GL or SSS modules result in slight performance degradation compared to the Baseline, while Ada-KD consistently recovers accuracy. Both GL+Ada-KD and SSS+Ada-KD surpass the Baseline across all metrics, confirming Ada-KD's role in maintaining detection performance under compression, particularly on PIDray's hard and hidden subsets. The complete GL+SSS+Ada-KD (GSA-YOLO) architecture achieves the best overall performance, with gains of 1.2\% in Precision, 1.7\% in Recall, 2.1\% in mAP50, and 1.8\% in mAP50:95 over the Baseline. 

\revise{To validate the effectiveness of our sigmoid-based adaptive decay schedule, we conducted comparative experiments with alternative decay strategies including cosine and linear schedules on the HiXray dataset. Results show that our complete GSA-YOLO model with sigmoid decay achieves Precision of 0.902, Recall of 0.795, mAP50 of 0.827, and mAP50:95 of 0.531, outperforming cosine decay at 0.526 and linear decay at 0.523, showing an improvement over cosine and linear decay schedules in this setting.} 
Table \ref{tab:ada_kd_comparison} presents the detailed comparative results.

\begin{table}[htbp]
\centering
\caption{\revise{Comparative analysis of Ada-KD decay schedules on HiXray dataset.}}\label{tab:ada_kd_comparison}
\resizebox{0.8\textwidth}{!}{%
\begin{tabular}{lcccc} 
\toprule
Method & Precision & Recall & mAP50 & mAP50:95 \\
\midrule
\multicolumn{5}{l}{\textit{Decay Schedule Comparison}} \\
\revise{Sigmoid Decay (Ours)} & \revise{\textbf{0.902}} & \revise{\textbf{0.795}} & \revise{\textbf{0.827}} & \revise{\textbf{0.531}} \\
\revise{Cosine Decay} & \revise{0.897} & \revise{0.789} & \revise{0.822} & \revise{0.526} \\
\revise{Linear Decay} & \revise{0.893} & \revise{0.785} & \revise{0.818} & \revise{0.523} \\
\bottomrule
\end{tabular}
}
\end{table} 

\revise{The detection performance across both datasets indicates that the integration of GL, SSS, and Ada-KD addresses X-ray detection challenges including severe occlusion, complex backgrounds, and real-time inference requirements}.

\subsection{Deployment and Training Cost Analysis}
\label{subsec:deployment training cost}

\newrevise{To evaluate GSA-YOLO's practical deployment readiness, we compare two inference configurations: the original PyTorch FP32 model and a TensorRT-optimized FP16 version.} As shown in Table~\ref{tab:deployment}, \newrevise{the TensorRT FP16 version reduces model size by approximately 49\% and improves inference throughput by approximately 70\%, with a modest accuracy drop of 0.013 in both mAP50 and mAP50:95. These results confirm that GSA-YOLO is compatible with standard inference optimization pipelines and suitable for lightweight deployment without significant accuracy degradation.}

\begin{table}[htbp]
\centering
\caption{\newrevise{Deployment configuration comparison for GSA-YOLO on HiXray.}}\label{tab:deployment}
\resizebox{0.85\textwidth}{!}{%
\begin{tabular}{llcccc}
\toprule
Model & Backend & Model Size & FPS & mAP50 & mAP50:95 \\
\midrule
\newrevise{GSA-YOLO} & \newrevise{PyTorch FP32} & \newrevise{5.7MB} & \newrevise{189.62} & \newrevise{0.827} & \newrevise{0.531} \\
\newrevise{GSA-YOLO} & \newrevise{TensorRT FP16} & \newrevise{2.9MB} & \newrevise{322.35} & \newrevise{0.814} & \newrevise{0.518} \\
\bottomrule
\end{tabular}
}
\end{table}

\newrevise{We further report the training cost of GSA-YOLO relative to representative baselines on HiXray.} As shown in Table~\ref{tab:training_cost}, \newrevise{GSA-YOLO requires approximately 2.1$\times$ more training time than YOLOv8n due to its three-stage pipeline. However, this is a one-time cost: at deployment, GSA-YOLO achieves the highest inference FPS (189.62) among all compared methods—3.5$\times$ faster than DSF-YOLO and 2.3$\times$ faster than GEMA-YOLO—while also achieving the best mAP50:95 (0.531) with the smallest model size (5.7 MB). The training overhead is thus justified by the superior accuracy-efficiency trade-off at inference time.}

\begin{table}[htbp]
\centering
\caption{\newrevise{Training cost and inference performance comparison on HiXray.}}\label{tab:training_cost}
\resizebox{0.9\textwidth}{!}{%
\begin{tabular}{lccccc}
\toprule
Model & Model Size & Training Time & Infer FPS & mAP50 & mAP50:95 \\
\midrule
\newrevise{YOLOv8n} & \newrevise{6.2MB} & \newrevise{5.23h} & \newrevise{170.58} & \newrevise{0.806} & \newrevise{0.507} \\
\newrevise{Fine-YOLO} & \newrevise{46.2MB} & \newrevise{5.35h} & \newrevise{152.42} & \newrevise{0.811} & \newrevise{0.514} \\
\newrevise{DSF-YOLO} & \newrevise{35.4MB} & \newrevise{5.78h} & \newrevise{54.60} & \newrevise{0.817} & \newrevise{0.515} \\
\newrevise{GEMA-YOLO} & \newrevise{11.8MB} & \newrevise{5.47h} & \newrevise{82.57} & \newrevise{0.822} & \newrevise{0.523} \\
\newrevise{GSA-YOLO (Ours)} & \newrevise{\textbf{5.7MB}} & \newrevise{10.15h} & \newrevise{\textbf{189.62}} & \newrevise{\textbf{0.827}} & \newrevise{\textbf{0.531}} \\
\bottomrule
\end{tabular}
}
\end{table}

\subsection{Visualization Analysis}
\label{subsec:visualization analysis}

\begin{figure}[htbp] 
    \centering
    \begin{minipage}[t]{0.48\textwidth}
        \centering
        \includegraphics[width=\textwidth]{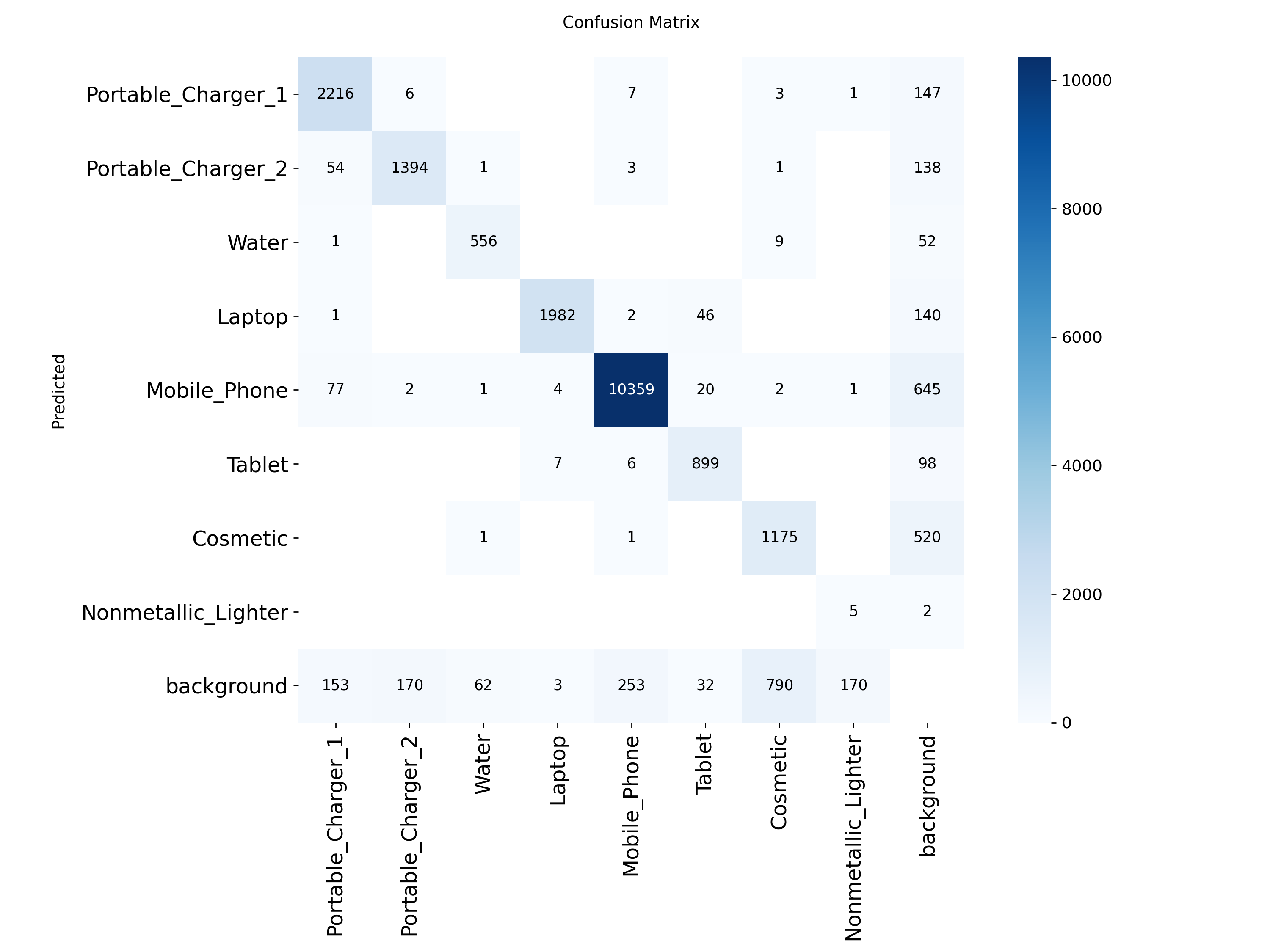} 
        \centerline{(a) HiXray: YOLOv8n Baseline} 
    \end{minipage}
    \hfill 
    \begin{minipage}[t]{0.48\textwidth}
        \centering
        \includegraphics[width=\textwidth]{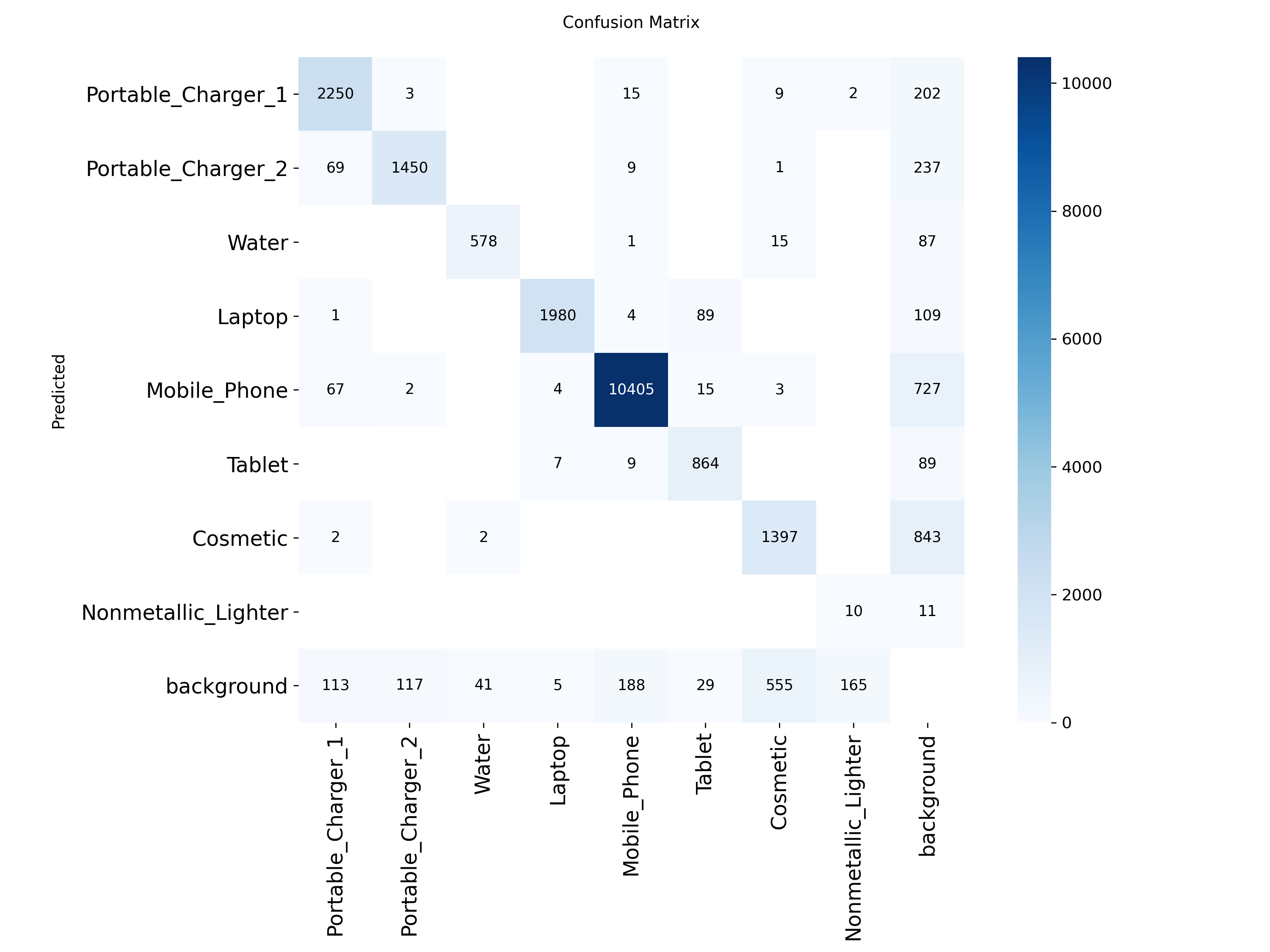} 
        \centerline{(b) HiXray: GSA-YOLO} 
    \end{minipage}
    
    \vspace{10pt} 
    
    \begin{minipage}[t]{0.48\textwidth}
        \centering
        \includegraphics[width=\textwidth]{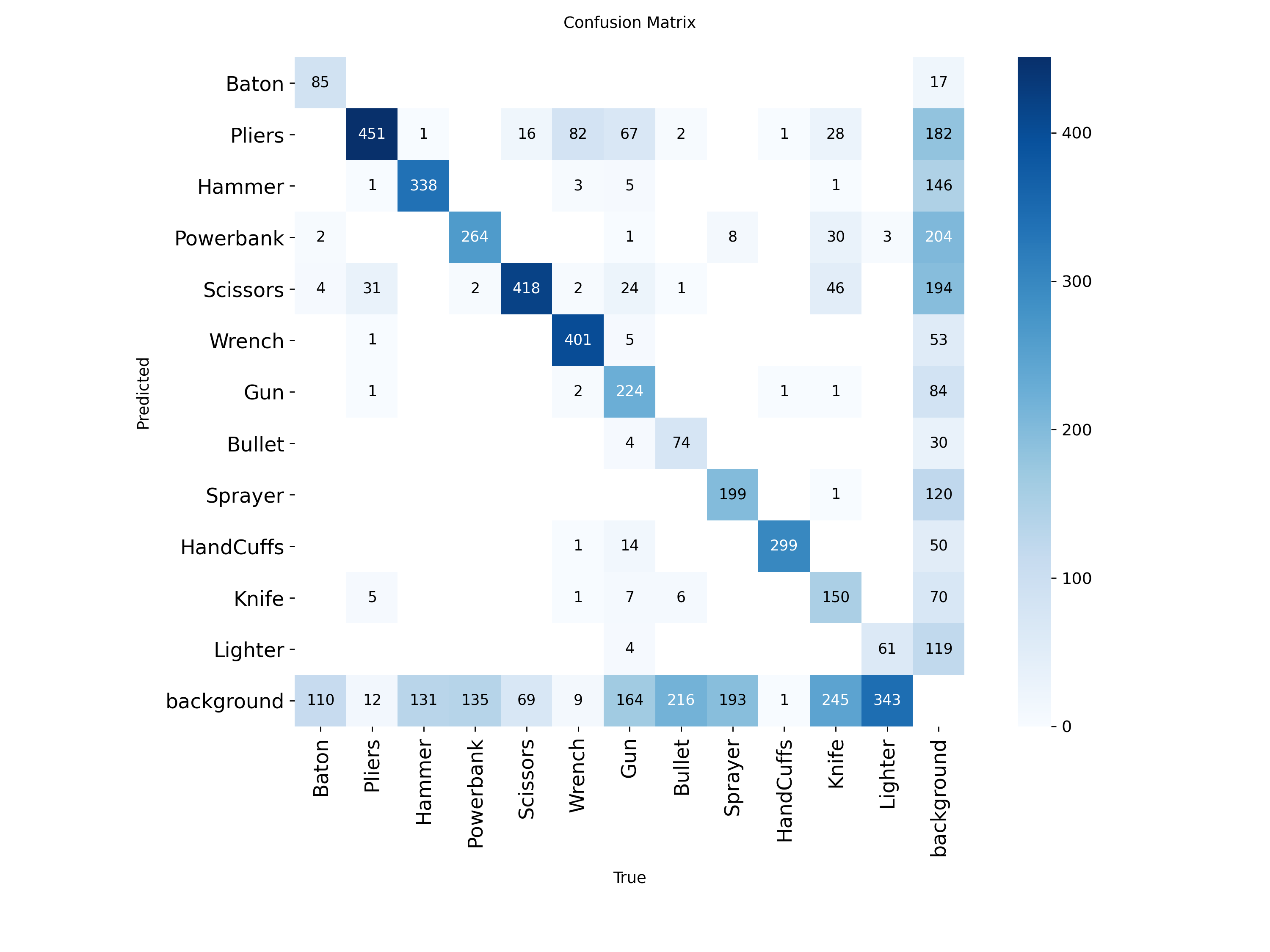} 
        \centerline{(c) PIDray(hidden): YOLOv8n Baseline} 
    \end{minipage}
    \hfill
    \begin{minipage}[t]{0.48\textwidth}
        \centering
        \includegraphics[width=\textwidth]{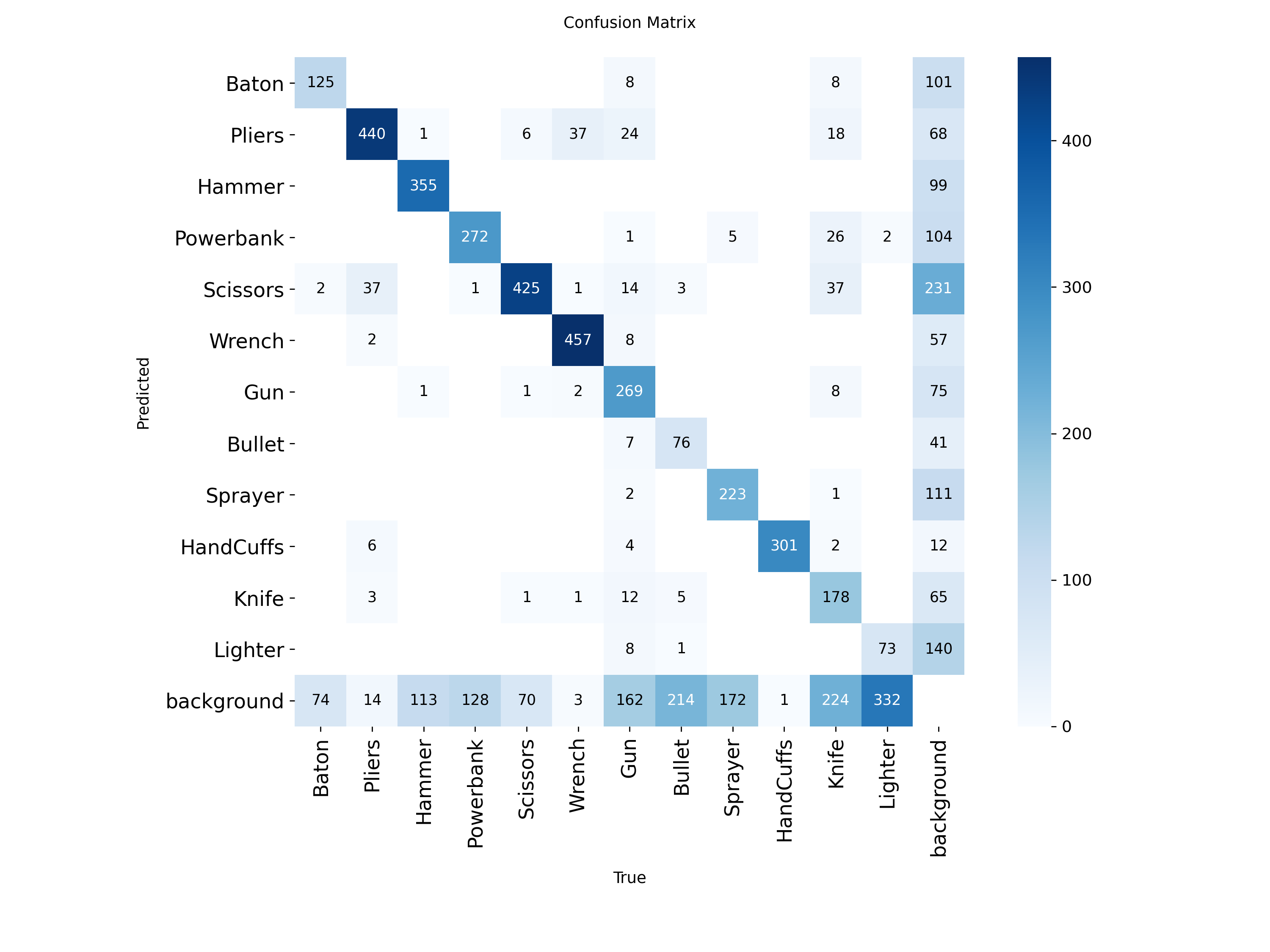} 
        \centerline{(d) PIDray(hidden): GSA-YOLO} 
    \end{minipage}
    
    \caption{Confusion matrix plots on the HiXray and PIDray hidden test sets. Subfigures (a) and (b) show the results on the HiXray test set for the YOLOv8n baseline and our GSA-YOLO model, respectively. Subfigures (c) and (d) show the corresponding results on the challenging PIDray hidden test set.}\label{fig:confusion_matrices}
\end{figure}

\revise{To examine the classification performance and detection completeness of GSA-YOLO, we visualize confusion matrices for the YOLOv8n baseline and our model on HiXray and PIDray hidden test sets, as shown in Figure} \ref{fig:confusion_matrices}. \revise{The analysis focuses on two performance indicators: True Positive (TP) counts on the diagonal, reflecting classification accuracy; and False Negative (FN) counts in the background row for non-background classes, representing missed detections}.

On the HiXray dataset, as shown in Figures \ref{fig:confusion_matrices}a and \ref{fig:confusion_matrices}b, \revise{GSA-YOLO shows generally higher TP counts and reduced missed detections across most categories}. For example, \revise{correctly identified Mobile\_Phone instances increased from 10,359 (baseline) to 10,405 (ours)}. \revise{GSA-YOLO reduces missed detections: for the Cosmetic category, missed objects decreased from 790 (baseline) to 555; missed detections for Portable\_Charger\_1 and Portable\_Charger\_2 decreased from 153 to 113 and from 170 to 117 respectively}. This pattern \revise{is observed on the PIDray hidden test set containing complex and occluded instances}. As shown in Figure \ref{fig:confusion_matrices}c and \ref{fig:confusion_matrices}d, \revise{GSA-YOLO increases correctly detected Wrench and Gun from 401 to 457 and 224 to 269 respectively}. \revise{These results indicate that GSA-YOLO maintains or improves detection completeness relative to the baseline across both datasets.}

\subsection{Feature Activation Analysis}
\label{subsec:feature activation analysis}

\begin{figure}[htbp]
\centering
\includegraphics[width=1\textwidth]{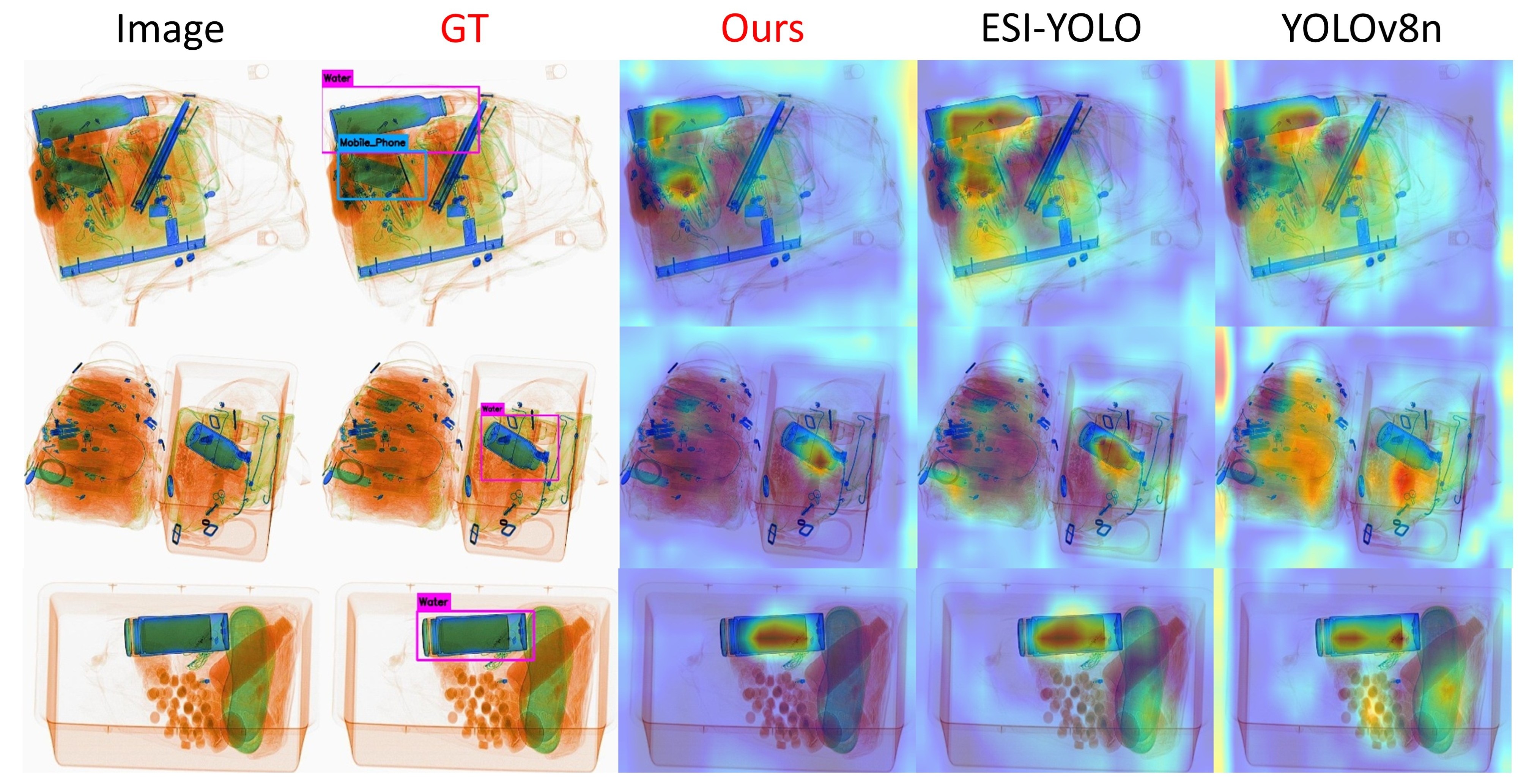}
\caption{\revise{Class Activation Mapping visualization comparison for X-ray prohibited item detection. The heat maps demonstrate the spatial attention patterns of different models: GSA-YOLO (Ours), ESI-YOLO, and YOLOv8n baseline, with Ground Truth (GT) annotations for reference.}}\label{fig:feature_activation}
\end{figure}

\revise{To examine the feature learning behavior of GSA-YOLO, we conduct Class Activation Mapping (CAM) visualization across three representative scenarios: multi-target occlusion (Water + Mobile\_Phone), concealed targets (liquid bottle in backpack), and clear single targets (isolated liquid bottle).} As shown in Figure \ref{fig:feature_activation}, \revise{GSA-YOLO produces more focused activation patterns with better alignment to target boundaries compared to ESI-YOLO and YOLOv8n, which exhibit broader or more diffuse activations. These observations are consistent with the expected effects of GL's soft regularization on feature representations and SSS's structured pruning on channel redundancy.}

\subsection{Visualization Analysis}
\label{subsec:visualization analysis2}
\begin{figure}[htbp]
    \centering
    \includegraphics[width=1\textwidth]{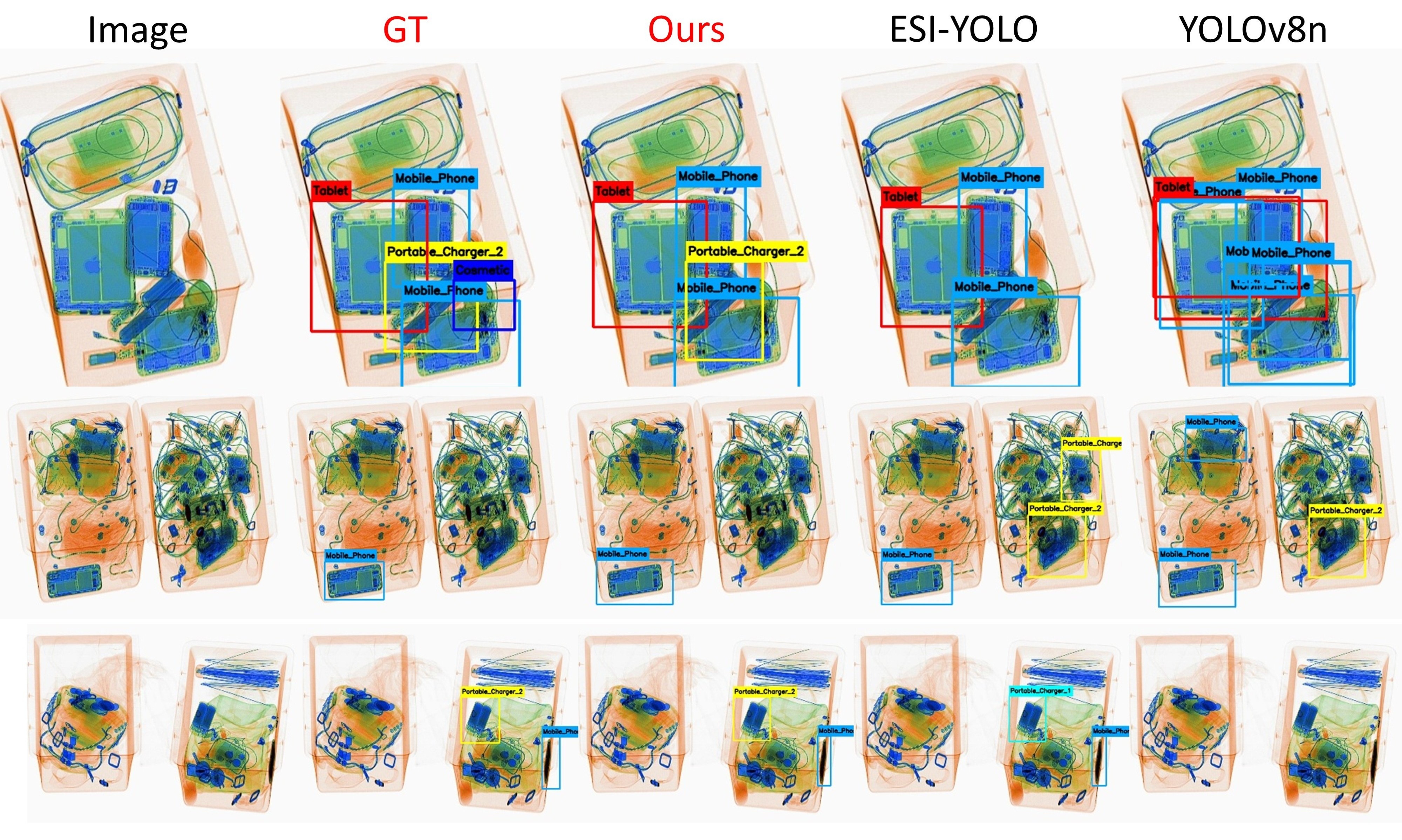}
    \caption{\revise{X-ray security inspection detection visualization comparison. The figure shows detection results across different complexity scenarios: Image (original X-ray images), GT (Ground Truth annotations), Ours (GSA-YOLO), ESI-YOLO, and YOLOv8n baseline. Different colored boxes represent prohibited items: red (Tablet), blue (Mobile Phone), yellow (Portable Charger), dark blue (Cosmetic). The comparison covers clear scenarios, complex occlusion scenarios, and overlapping small target scenarios.}}
    \label{fig:detection_comparison}
\end{figure}

\revise{To demonstrate the detection effectiveness in real X-ray security inspection scenarios, we conducted visual comparison among three models: YOLOv8n baseline, ESI-YOLO, and GSA-YOLO. We selected representative X-ray images covering different complexity levels including clear scenarios, complex occlusion scenarios, and overlapping small target scenarios. The comparative detection results are presented in Figure} \ref{fig:detection_comparison}. 

\revise{The results are presented across three complexity levels. In clear scenarios (first row), GSA-YOLO produces bounding boxes that are well-aligned with Ground Truth, while YOLOv8n shows duplicate detections and positional offsets. In complex occlusion scenarios (second row), GSA-YOLO maintains stable detection for items obscured by cables, whereas YOLOv8n exhibits more missed detections. In overlapping small target scenarios (third row), GSA-YOLO localizes overlapping portable chargers and mobile phones more accurately, while YOLOv8n shows larger positional deviations for small targets. ESI-YOLO performs comparably to GSA-YOLO in most scenarios but shows slight disadvantages in small target recall}.

\section{Conclusion}
\label{sec:conclusion}
In this paper, we introduced GSA-YOLO, a lightweight framework designed for efficient and robust prohibited item detection in X-ray security imagery. To address the need for real-time models resilient to severe occlusion and complex clutter, GSA-YOLO integrates three core components: Group Lasso (GL) for feature refinement, Sparse Structure Selection (SSS) for structured pruning, and Adaptive Knowledge Distillation (Ada-KD) for accuracy recovery. Experiments on the HiXray and PIDray datasets confirm the framework's overall effectiveness. GSA-YOLO successfully reduced the computational cost from the YOLOv8n baseline’s 8.7G to 8.0G, while simultaneously achieving an mAP50:95 of 0.531 on HiXray, a 2.4\% improvement over the baseline, and 0.679 on PIDray, a 1.8\% lift, \revise{demonstrating its potential as a promising candidate for efficient X-ray security inspection under controlled benchmark conditions.}

Despite these results, the GSA-YOLO framework has certain limitations. The current implementation relies on a sequential, multi-stage optimization process, GL $\rightarrow$ SSS $\rightarrow$ Ada-KD. Due to the non-differentiable nature of hard pruning, this pipeline increases sensitivity to hyperparameter tuning and requires careful calibration across stages. \newrevise{Furthermore, X-ray prohibited item detection relies on fine-grained visual cues—such as overlapping objects and low-contrast materials—that are sensitive to aggressive channel pruning; this domain-specific constraint motivates the adoption of low-to-moderate compression, as verified by the experimental comparisons against stronger compression baselines.} \revise{Regarding practical deployment feasibility, the pruned GSA-YOLO model achieves a model size of 5.7 MB and 8.0 GFLOPs, both lower than the YOLOv8n baseline (6.2 MB, 8.7 GFLOPs), indicating a compact memory footprint compatible with edge inference platforms; however, comprehensive evaluation of latency fluctuations under variable input conditions and hardware compatibility with proprietary security inspection pipelines remains an important direction for future work.} Future work will focus on two main directions: first, investigating GSA-YOLO's deployment on resource-limited edge devices using mixed-precision quantization; and second, extending the GSA components to other advanced object detection architectures beyond the YOLO family. \revise{Additionally, evaluating the multi-stage training pipeline under alternative optimizers such as AdaBoB}~\cite{xiang_adabob_2025}\revise{, which combines adaptive gradient confidence and dynamic learning rate boundaries, represents a promising direction for further improving training stability and convergence.} \newrevise{Beyond optimizer choice, recent metaheuristic methods such as the Secant Optimization Algorithm (SOA)} \cite{ibrahim_soa_2026} \newrevise{and the Schr\"{o}dinger Optimizer (SRA)} \cite{hussein_sra_2025} \newrevise{offer complementary potential: their adaptive exploration-exploitation balance is conceptually aligned with the dynamic weighting in Ada-KD, making them promising candidates for automated hyperparameter search in the GSA-YOLO pipeline.}

\section*{Declaration of conflicting interest}
The authors declare no conflicts of interest.

\section*{Funding}
The authors did not receive support from any organization for the submitted work.

\section*{Ethical approval and informed consent statements}
Not applicable.

\section*{Data availability statement}
The datasets used and/or analysed during the current study are available from the corresponding author on reasonable request.

\section*{Authors' contributions}
J.K. conceived and designed the study, performed data collection and statistical analysis, interpreted the results, and wrote the main manuscript text. J.K. also prepared all tables and figures. The author reviewed and approved the final manuscript.

\bibliographystyle{elsarticle-num}
\bibliography{related_work}

\end{document}